\documentclass[sigconf, screen]{acmart}

\AtBeginDocument{%
  \providecommand\BibTeX{{%
    \normalfont B\kern-0.5em{\scshape i\kern-0.25em b}\kern-0.8em\TeX}}}

\copyrightyear{2023}
\acmYear{2023}
\setcopyright{acmlicensed}\acmConference[WWW '23 Companion]{Companion
Proceedings of the ACM Web Conference 2023}{April 30-May 4, 2023}{Austin,
TX, USA}
\acmBooktitle{Companion Proceedings of the ACM Web Conference 2023 (WWW
'23 Companion), April 30-May 4, 2023, Austin, TX, USA}
\acmPrice{15.00}
\acmDOI{10.1145/3543873.3587669}
\acmISBN{978-1-4503-9419-2/23/04}

\usepackage[utf8]{inputenc} 
\usepackage[T1]{fontenc}    
\usepackage{hyperref}       
\usepackage{url}            
\usepackage{booktabs}       
\usepackage{amsfonts}       
\usepackage{nicefrac}       
\usepackage{microtype}      
\usepackage{graphicx}
\usepackage{amsmath,amsfonts,amsthm}
\usepackage{array}
\usepackage{subcaption}
\usepackage{bbm}
\usepackage{bm}
\usepackage{multicol}
\usepackage{lipsum}
\usepackage{wrapfig}
\usepackage[ruled]{algorithm2e}
\usepackage{arydshln}
\usepackage{multirow}
\usepackage{enumitem}
\usepackage{footmisc}

\newtheorem{proposition}{Proposition}

\newtheorem{remark}{Remark}
\newtheorem{theorem}{Theorem}

\newtheorem{example}{Example}

\newcommand{\Ical}{\mathcal{I}}

\newcommand{\Fcal}{\mathcal{F}}
\newcommand{\Gcal}{\mathcal{G}}
\newcommand{\Xcal}{\mathcal{X}}
\newcommand{\Ecal}{\mathcal{E}}
\newcommand{\Rcal}{\mathcal{R}}
\newcommand{\Ocal}{\mathcal{O}}

\newcommand{\Ebb}{\mathbb{E}}
\newcommand{\Rbb}{\mathbb{R}}

\newcommand{\ybf}{\mathbf{y}}

\newcommand{\Phibf}{\mathbf{\Phi}}

\begin{document}

\title{Pretrained Embeddings for E-commerce Machine Learning: When it Fails and Why?}

\author{Da Xu}
\authornote{The work was done when the authors was with Walmart Labs.}
\affiliation{%
  \institution{LinkedIn}
  \city{Sunnyvale}
  \state{California}
  \country{USA}
}
\email{daxu5180@gmail.com}

\author{Bo Yang}
\affiliation{%
  \institution{Amazon}
  \city{Sunnyvale}
  \state{California}
  \country{USA}
}
\email{boyang.emma@gmail.com}




\begin{abstract}
The use of pretrained embeddings has become widespread in modern e-commerce machine learning (ML) systems.
In practice, however, we have encountered several key issues when using pretrained embedding in a real-world production system, many of which cannot be fully explained by current knowledge.
Unfortunately, we find that there is a lack of a thorough understanding of how pre-trained embeddings work, especially their intrinsic properties and interactions with downstream tasks.
Consequently, it becomes challenging to make interactive and scalable decisions regarding the use of pre-trained embeddings in practice.



Our investigation leads to two significant discoveries about using pretrained embeddings in e-commerce applications. 
Firstly, we find that the design of the pretraining and downstream models, particularly how they encode and decode information via embedding vectors, can have a profound impact.
Secondly, we establish a principled perspective of pre-trained embeddings via the lens of kernel analysis, which can be used to evaluate their predictability, interactively and scalably. 
These findings help to address the practical challenges we faced and offer valuable guidance for successful adoption of pretrained embeddings in real-world production. 
Our conclusions are backed by solid theoretical reasoning, benchmark experiments, as well as online testings.


\end{abstract}




\maketitle

\section{Introduction}
\label{sec:introduction}
As deep learning continues to reshape the landscape of machine learning, pretrained embeddings have emerged as a crucial component in e-commerce ML systems. Researchers are continuously working to improve the design of pre-training algorithms for more complex data structures, \cite{barkan2016item2vec,wang2018billion,xu2020product,vasile2016meta,wan2018representing,xu2020knowledge,zhang2020towards}, while practitioners have made substantial strides in optimizing and deploying embeddings for large-scale industrial applications \cite{covington2016deep,huang2020embedding,grbovic2018real}.

For e-commerce ML, the use for pretrained embedding is driven by the sparse and unstructured nature of data, such as user behavior, feedback, catalogs, reviews, texts, images, etc. 
Therefore, deep learning's ability to automatically learn vector representations of data is highly appealing compared to traditional feature engineering methods, as it innovates the \emph{lifecycle} and \emph{workflow} of ML systems.  
Downstream model owners can simply query the \emph{embedding store}, eliminating the need to construct features from scratch.

In recent years, embedding-based solutions are gradually dominating e-commerce ML, with two of the most important application areas being \emph{recommendation} \cite{fang2020deep,xu2020knowledge,fei2019end,saveski2014item} and \emph{content understanding}, including tasks such as \emph{knowledge (graph) learning} \cite{xu2020product,li2020alimekg,dong2018challenges}, \emph{user and item segmentation} \cite{luo2019conceptualize,xu2021theoretical,gu2020hierarchical,wan2018representing,jie2021bidding}.
Despite some investigations have attemtped to justify pretrained embeddings \cite{allen2019analogies,arora2016latent,xu2021theoretical}, there is no one-size-fits-all approach to pretraining or using embeddings for these tasks, and the manner in which pretrained embeddings encode and decode information can vary greatly.
As a result, the efficacy of pretrained embeddings is often determined through empirical evaluation rather than solid theoretical understanding.

\textbf{What went wrong in our production?} In our e-commerce platform, we experienced first-hand that using pretrained embeddings for downstream tasks can be highly non-trivial. In particular, two of the biggest challenges were:
\begin{itemize}[leftmargin=*]
    \item pretrained embeddings can cause \textbf{instability} issue, that is, the downstream performance is unstable even if the pretraining data and configurations are kept the same;
    \item a lack of method to examine the \textbf{predictability} of pretrained embeddings for the downstream tasks, e.g. existing metrics do not attribute the downstream performance to embedding accurately. 
\end{itemize}
If left unaddressed, these two issues could pose risks to the \emph{reliability} of industrial production systems. Specifically, it is difficult for developer to make \textbf{interactive} and \textbf{scalable} decisions regarding the design of pretraining and the quality of pretrained embeddings, because the examination via empirical performance requires running the entire pretraining and downstream training pipeline.

To help the readers understand the reality of these challenges, we use the \emph{contrastive representation learning} as an example. Let $\phi(x_1)$ and $\phi(x_2)$ be the two "ground truth" embedding vectors for \emph{entity} $x_1$ and $x_2$. Recall that the most common method for pretraining, stochastic gradient descent (SGD), can only converge (stochastically) to a local region of the ground truth \cite{nocedal1999numerical}. It means that instead of getting $\phi(x_1)$ and $\phi(x_2)$, we are more likely to obtain $\Phi(x_1)$ and $\Phi(x_2)$ who are random variables that satisfy $\Ebb \Phi(x_1) = \phi(x_1)$ and $\Ebb \Phi(x_2) = \phi(x_2)$. Since the initializations are coordinate-wise $i.i.d$, it indicates that $\Phi(x_1)$ and $\Phi(x_2)$ also have independent coordinates. As a result, $\big\langle \phi(x_1) , \phi(x_2) \big\rangle$ can be a good approximation of $\Ebb \big[\big\langle \Phi(x_1) , \Phi(x_2) \big\rangle \big]$, which implies that even if the pretraining model converges to \emph{deterministic output}, the pretrained embeddings can still retain a degree of \emph{randomness}. In Section \ref{sec:preliminary}, we elaborate this phenomenon using several benchmark datasets.
If downstream developers apply $\Phi(x)$ instead of $\phi(x)$, it can be anticipated that they experience \emph{stability} and \emph{predictability} issues in the downstream performance. 


As a matter of fact, when the pretraining algorithm \textbf{encodes} information using a particular function (e.g. inner product), it implicitly restricts how the information should be accessed. Therefore, when the downstream models attempt to \textbf{decode} the information using other functions, they may suffer from additional noise. 
The \textbf{first contribution} of this paper is to formally characterize the consequences of \emph{mismatch between the encoding and decoding functions} for contrastive representation learning. In Section \ref{sec:preliminary} and \ref{sec:structure}, we provide both numerical and theoretical results to elaborate our findings. We conclude that using a matching pretraining-downstream design is crucial for the \emph{stability} of downstream performance.

The second contribution of this paper is the development of \emph{metrics} to evaluate the predictability of pretrained embeddings in e-commerce applications, especially for entity-wise and sequential tasks. The metrics are designed to not only retain the pretraining information but also to predict the downstream performances. This work is inspired by recent progress in deep learning theory, specifically the examination of deep learning models through the lens of \textbf{kernel} (Section \ref{sec:kernel}).

The \textbf{third contribution} of our paper to corroborate our analysis and findings with offline experiments and real-world online testings. We also provide reproducible benchmark experiments for the benefit of researchers and practitioners to further explore the challenges and improve the use of pretrained embeddings in industrial machine learning systems.


\section{Preliminaries}
\label{sec:preliminary}
We use $x\in\Xcal$ to denote the \emph{entities} such as \emph{items} and \emph{users}. We now let $\phi: \Xcal \to \Rbb^d$ to be the embedding mapping, which can be parameterized by a native vector or complex neural networks. We use $d$ to denote the embedding dimension.
The downstream task uses a prediction function $f\in\Fcal$ to produce the score for the machine learning tasks.
We mainly consider two tasks, one is \textbf{entity classification} that is crucial for many user and item segmentation tasks, the other is \textbf{sequential recommendation} that leverages the past interacted items to make personalized, intent-aware recommendations.

\subsection{Data and Illustration Experiments}

We begin by outlining the settings of our illustration experiments because they will appear frequently in the subsequent sections. All the public implementation code are provided in \textbf{online material} \footnote{\url{https://github.com/daxu5180/WebConf23}}. All the reported standard deviations are computed from ten independent runs. 
The datasets for our experiments are summarized as below:
\begin{itemize}[leftmargin=*]
    \item \textbf{Instacart}: the Instacart data collect the online grocery baskets, and consists of $\sim$3 million orders from around 50,000 products and 200,000 users. Each grocery item has the title information and the aisle (category) of the item.
    \item \textbf{Amazon Electronics}: the data consists of $\sim$1 million reviews from the customers who have purchased electronic products from Amazon, collected from $\sim$200 thousand users with around 60,000 items. The metadata includes both the item titles and their category hierarchy.
    \item \textbf{MovieLens-1M}\footnote{Although MovieLens is not a e-commerce dataset, it contains metadata that are useful for elaborating our analysis.}: the dataset collects the users' ratings of the movies. It consists of around 1 million ratings from 6,040 users on 3,925 movies. The dataset has been preprocessed, and we covert the rating to implicit feedback (interacted or not). Each movie has the title information and the genre categorization.
    \item \textbf{Deployment platform of \textbf{'ECOM'}}: we have the access to all the online shopping data of 'ECOM' -- a major e-commerce platform in the U.S hosting \emph{hundreds of millions} of grocery and general merchandise products. Our online deployments and A/B testings are implemented on the production environment of 'ECOM'.
\end{itemize}
By convention, we remove items and users who have less than \emph{five} total interactions. For the benchmark experiments, we employ the bag-of-items \emph{Item2vec} \cite{barkan2016item2vec} and \emph{BERT} \cite{devlin2018bert} as the embedding pretraining models. 
They possess different \emph{encoding structures} and are suitable different pretraining data: 
\begin{itemize}[leftmargin=*]
    \item \textbf{Item2vec} follows the same unsupervised representation learning approach based on a bag-of-items as \emph{Word2vec}. It utilizes user behavior data (e.g. sequences of views or purchases) to derive \emph{functional similarity} among items and encode it using the inner product of embeddings.
    \item \textbf{BERT} is an advanced NLP model that employs a sophisticated architecture to extract and encode textual data related to an item (e.g. title and description). Typically, the last hidden layer of BERT is used as the item embedding, and a linear model can be used to access the encoded embedding information.
\end{itemize}


For the downstream \textbf{entity classification} task, we utilize the provided item catalog information as the label, such as the \emph{shelf}, \emph{department}, and \emph{genre}, which are provided in the metadata. To evaluate the downstream performance, we use a random 80\%-10\%-10\% split of the items and compute the \textbf{Micro-F1} and \textbf{Macro-F1} scores on testing data. For the downstream \emph{content understanding} models, we use will the standard \emph{logistic regression} (LR) and \emph{inner-product kernel support vector machine} (IP), as they represent two common methods for decoding information from pretrained embeddings.
\begin{itemize}[leftmargin=*]
    \item \emph{logistic regression} (\textbf{\emph{LR}}) directly applies a \emph{linear} layer which can be viewed as a simplified \emph{feed-forward} structure;
    \item \emph{inner-product kernel support vector machine} (\textbf{\emph{IP}}) first creates the kernel by computing the pairwise inner product of the pretrained embeddings and them apply SVM. Its essence is to use \emph{inner product} for decoding.
\end{itemize}

For the downstream \textbf{sequential recommendation} task, we adopt the standard leave-last-one-out training, validation and evaluation setup introduced in \cite{kang2018self}. We use the top-10 \textbf{recall} (\emph{Hit@10}) and mean reciprocal rank (\emph{MRR@10}), and the \emph{overall NDCG} as evaluation metrics. The downstream models are given by the \emph{MLP4Rec}, \emph{GRU4Rec}, and \emph{attention-based} recommendation (\emph{Attn4rec}) \cite{zhang2019deep}, each representing a prevalent class of sequential recommendation model. We defer their detailed introductions to Appendix \ref{append:experiment}.




\subsection{Pretraining-downstream Configurations}

In e-commerce ML,  \emph{user behavior} data, which refers to how users interact with items, and \emph{item catalog} data, which includes contextual and descriptive information about the products, are the two primary sources of pretraining information. In some rare cases, user demographic features may also be available, but we do not discuss this scenario due to privacy concerns.

Learning embeddings from customer behavior is often accomplished using \emph{unsupervised} techniques, among which \textbf{{contrastive representation learning (CL)}} is a widely used approach. Generally, CL requires that the \emph{inner product} of "semantically similar" item representations is larger than that of randomly sampled negative pairs. We use $\Ical = \big\{i_1,\ldots,i_p\big\}$ to denote the set of items. For an item $i$, we denote the positive pair as $\big(i,i^+\big)$ and the negative pair as $\big(i,i^-\big)$. 
For clarity, we consider using one negative pair for each positive pair, so the objective function is given by:
\[
\Ebb \Big[ -\log \frac{e^{\phi(i)^{\intercal}\phi(i^+)}}{e^{\phi(i)^{\intercal}\phi(i^+)} + e^{\phi(i)^{\intercal}\phi(i^-)}} \Big],
\]
where the expectation is taken with respect to the underlying distribution of $(i,i^+,i^-)$. The \emph{Item2vec} model in our illustrative experiments belongs to this category.

One the other hand, as represented by \emph{BERT}, the pretraining model can be composed of a representation mapping and the subsequent feed-forward function: $\Gcal = \big\{f\circ\phi \,|\, f\in\Fcal, \phi \in \Phi\big\}$, where $\Fcal$ and $\Phi$ denote the respective function classes and $\circ$ denotes function composition. In what follows, we use $\hat{\phi}$ to denote the pretrained embedding.

Downstream ML models in e-commerce ML can also be categorized based on how they consume $\hat{\phi}$: 
\begin{itemize}[leftmargin=*]
    \item taking the inner product of $\hat{\phi}$ for such as \emph{K-nearest neighbor} or \emph{inner-product kernel support vector machine} (\textbf{IP}) methods;
    \item apply another feed-forward structure on $\hat{\phi}$, such as using {logistic regression} (\textbf{LR}) with $\hat{\phi}$ as the feature vector.
\end{itemize}

They give rise to two types of pretraining-downstream configurations: one use the same encoding-decoding structure, and the other use different encoding-decoding structures. For instance, the BERT-LR configuration is \textbf{homogeneous} because they both apply feed-forward functions on embeddings, meanwhile the CL-LR configuration is \textbf{heterogeneous}.

\subsection{The Instability Issue}

\begin{figure}[htb]
    \centering
    \includegraphics[width=\linewidth]{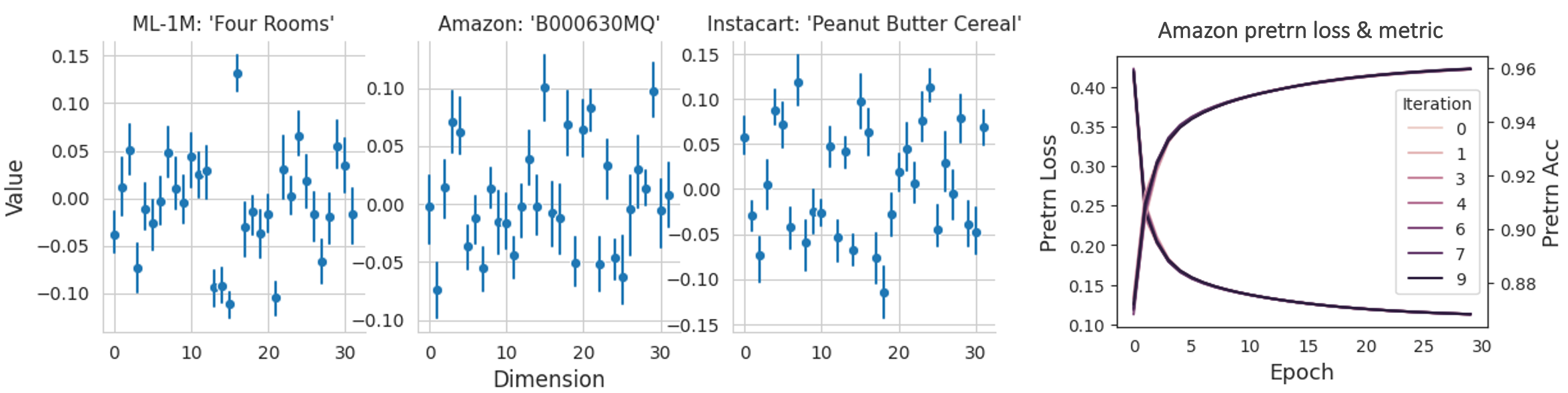}
    \caption{\small For each dataset, we randomly sample one item (movie) and visualize its embedding (with $d=32$ and normalized) obtained from CL-based (Item2vec) pretraining \cite{barkan2016item2vec}. The standard deviation bar is obtained from ten independent repetitions. To the right is the trajectory of the pretraining loss and accuracy during each independent repetition (with Amazon electronics as an example).}
    \label{fig:emb-var}
    \vspace{-0.15cm}
\end{figure}

In this section, we replicate our production issues with benchmark datasets to give readers a preliminary understanding of how pretrained embedding can lead to problems in a simple downstream task. In Figure \ref{fig:emb-var}, we first elaborate the \textbf{instability} issues associated with pretrained embeddings. 

We plot the pretrained embeddings from ten separate independent runs, where the same data, modelling, and optimization configurations are applied to each run (see Appendix \ref{append:experiment} for experiment setup). We can see that most entries in the embedding vectors exhibit \textbf{significant fluctuations}, even if the training loss and metrics almost follow the same trajectory in each run. Because we are using \emph{Item2vec} which is CL-based, the results empirically demonstrate our intuition in Section \ref{sec:introduction} that the pretraining algorithm actually produces a stochastic $\Phi(x)$ rather than the ideal $\phi(x)$.

\begin{figure}[htb]
    \centering
    \includegraphics[width=0.8\linewidth]{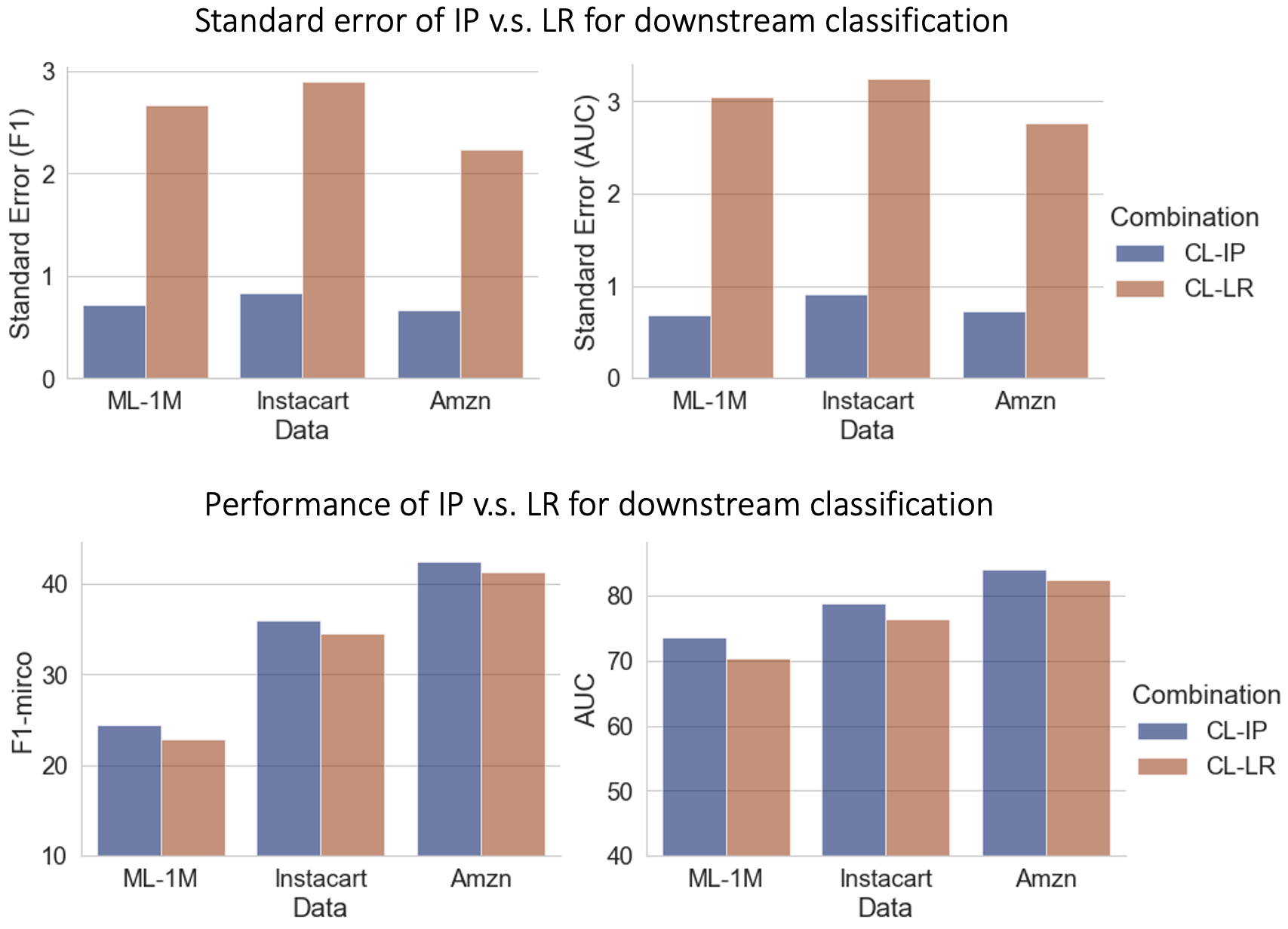}
    \caption{\small Upper: the standard errors (SE) of the testing performances (from ten independent repetitions); Lower: the testing performances. The results are from using \emph{IP} and \emph{LR} with CL-pretrained (\emph{Item2vec}) embeddings for downstream item classification. The results are multiplied by 100.}
    \label{fig:SD-CL}
\end{figure}

In the downstream task where we classify the items using the pretrained embeddings (upper panel of Figure \ref{fig:SD-CL}), we see that the classification outcomes are suboptimal and have high variances when LR is used as the downstream model, demonstrating that the \emph{heterogeneous configuration} can cause severe instability problems. On the other hand, Figure \ref{fig:SD-CL}) shows that the \emph{homogeneous configuration} (where IP is used as the downstream model in this case) gives much improved performance and stability.

In practice, \textbf{structure mismatch} is often overlooked due to belief that pretrained embeddings can be accessed without restrictions. In the next section, we formalize the problem and rigorously reveal the downstream performance gap between homogeneous and heterogeneous configurations.

\section{Impact of pretraining-downstream model structures}
\label{sec:structure}
To further investigate our previous observation, we can do a controlled experiment by generating the pretraining data in a particular manner. As before, the downstream task is item classification. During the CL pretraining, for each item $x$, we now sample $x^+$ from the same category (or genre), and $x^-$ is still drawn randomly from other categories. In this way, the distribution of $(x, x^+,x^-)$ is exactly the same as the downstream item classification task, which \textbf{eliminates} the potential confounding of distribution mismatch.

In this way, the data distributions are now the same for pretraining and the downstream item classification task. Therefore, any unexpected phenomenon in the downstream performance must be due to the models.
We use \textbf{CL-LR}, \textbf{CL-IP}, \textbf{BERT-IP}, and \textbf{BERT-IP} to denote the four pretraining-downstream configurations, where LR is logistic regression, and IP is inner-product kernel SVM. 

\begin{figure}[hbt]
    \centering
    \includegraphics[width=0.8\linewidth]{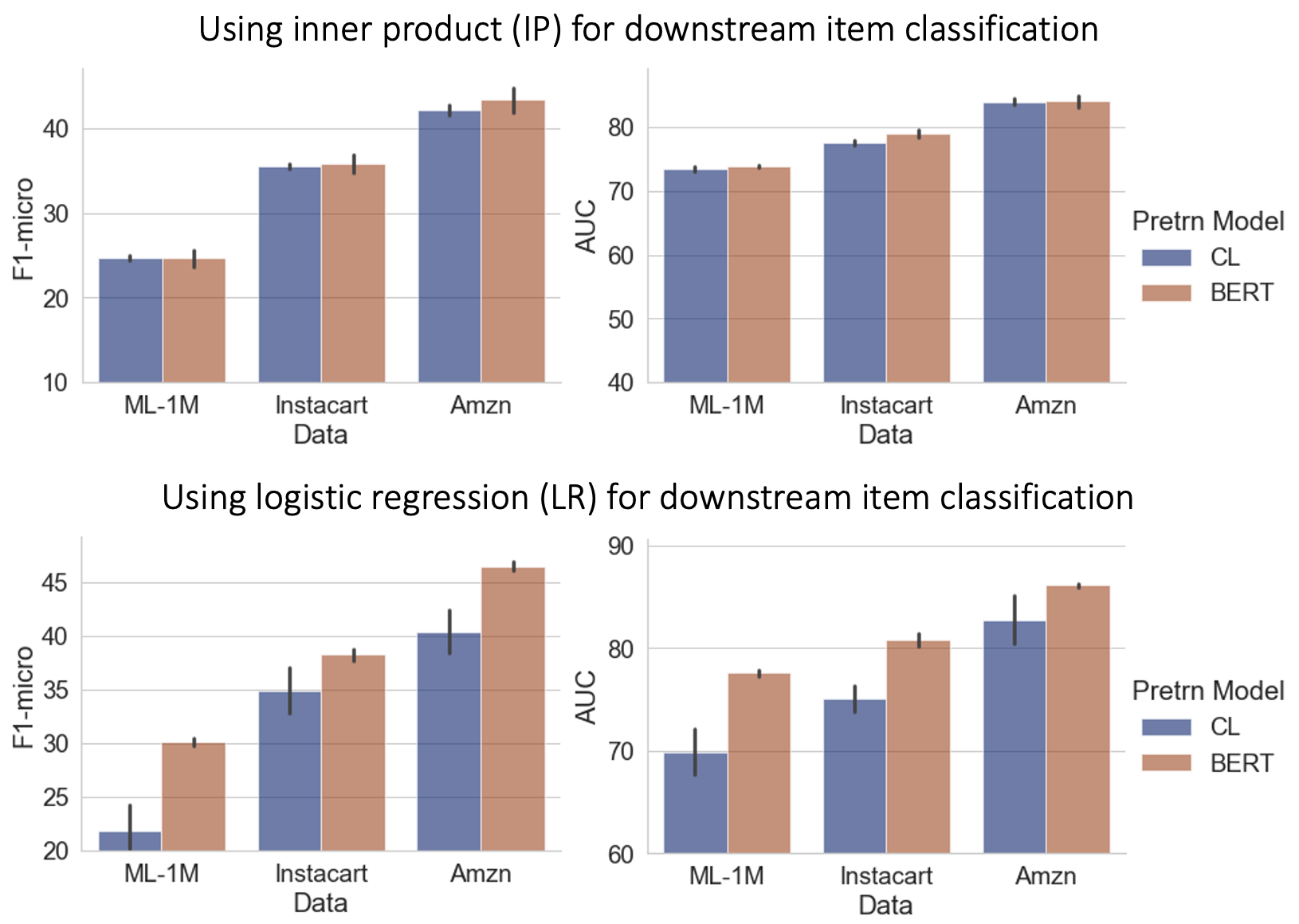}
    \caption{\small The impact of model design for i.i.d downstream task. Recall that BERT-LR, CL-IP represent the homogeneous configuration, or otherwise the pretraining and downstream structures are different. All the results are multiplied by 100. The details for he experiment setup are provided in Appendix \ref{append:experiment}.}
    \label{fig:LR-IP}
    \vspace{-0.1cm}
\end{figure}

In Figure \ref{fig:LR-IP}, we present the downstream testing result of our controlled pretraining setting. It can be observed that the best performance is achieved by the \emph{homogeneous configurations}, namely \textbf{CL-IP} and \textbf{BERT-LR}, for both CL- and BERT-based pretraining. On the other hand, the \emph{heterogeneous configurations} \textbf{CL-LR} and \textbf{BERT-IP} tend to have lower performance than their counterparts. 
Since the data mismatch factor has been eliminated, we hypothesize that the way pretraining and downstream models encode and decode embeddings has a significant impact on the downstream model's testing performance. In the following sections, we analyze this phenomenon using the framework of \textbf{generalization error bound} from statistical learning theory \cite{bartlett2002rademacher}.

\textbf{Theoretical justification}. 
To simplify the discussion, we assume that the downstream task involves binary classification using pretrained item embeddings. Consider \emph{CL-LR} and \emph{BERT-LR} as the two representatives for the heterogeneous and homogeneous configurations. In our analysis, BERT can actually be replaced by any model that has a feed-forward final layer. 

Let us begin by considering how the bag-of-items approach handles positive and negative pairs, its \emph{sampling distribution} $P_{\text{CL}}$ can be broken down into: 
\[
P_{\text{CL}}\big(x,x^+,x^-\big) = P_{\text{pos}}\big(x,x^+\big) P_{\text{neg}}\big(x^-\big).
\]
When denoting the supervised distribution associated with the labeling mechanism $\tau:\Xcal\to \{0,1\}$ as $P_{\tau}$, the downstream risk for a downstream model $f\circ\hat{\phi}$ can be expressed as follows:
\begin{equation*}
    \Ebb_{(x,y)\sim P_{\tau}} \ell \big( f\circ\hat{\phi}(x), y \big).
\end{equation*}
We highlight that the reason why CL often works well in practice can be attributed to the similarity between the distributions $P_{\text{pos}}$ and $P_{\text{neg}}$ and the supervised distribution $P_{\tau}$, in the sense that $\tau(x) = \tau\big(x^+\big)$ and $\tau(x) \neq \tau\big(x^-\big)$. To provide a formal framework for this observation, we introduce the concept of an \emph{environment} that controls both pretraining and downstream tasks:
\[\Ecal = \big\{\tau, P_X^{(0)}, P_X^{(1)}, P_Y\big\},\] 
where $\tau: \Xcal \to \{0,1\}$ is the downstream \emph{labelling mechanism}, $P_x^{(y)}$ is the conditional distribution $p(X=x | Y=y)$, and $P_Y$ is the labels' marginal distribution. With a slight abuse of notation, we use $P_{\tau} \sim \Ecal$ to denote the training distribution that is drawn from the environment $\Ecal$.


For any representation mapping $\phi\in\Phi$, the CL-based pretraining risk can be decomposed via: \[
R_{\text{CL}}(\phi) := R^C_{\text{CL}}(\phi) + R^I_{\text{CL}}(\phi),
\]
where $R^C_{\text{CL}}(\phi)$ corresponds to the risk under \emph{correct negative samples} where $x^-$ belongs to the class different from $x$ and $x^+$, and $R^I_{\text{CL}}(\phi)$ corresponds to the risk taken with respect to \emph{incorrect negative samples}. Again, this decomposition aims to eliminate the potential impact of distribution mismatch.

Since $R^C_{\text{CL}}(\phi)$ is risk associated with the "correct" triplets drawn from the same environment as the downstream task, the optimal risk that can possibly be achieved by CL is given by: 
\[R^*_{\text{pre}}:=\min_{\phi\in\Phi} R^C_{\text{CL}}(\phi).\] 

Moving on to the BERT-based setting, its two-stage optimization process can be described as follows:
\begin{enumerate}[leftmargin=*]
    \item the pretraining algorithm returns the embeddings $\hat{\phi}$;
    \item after plugging in $\hat{\phi}$, the downstream classifier is then optimized on the $n$ samples drawn from $P_{\tau}$, which we denote by: $f_{\hat{\phi},n}$.
\end{enumerate}
Therefore, the risk of the downstream task given $\hat{\phi}$ is: 
\begin{equation}
\label{eqn:phi-risk}
R_{\text{task}}^*(\hat{\phi}) = \Ebb_{P_{\tau}\sim \Ecal} \Ebb_{(x,y)\sim P_{\tau}} \ell\big(f_{\hat{\phi},n}(x), y \big),
\end{equation}
and again we let the pretraining and downstream samples both be drawn from the same environment to eliminate the confounding of distribution mismatch.

For both homogeneous and heterogeneous configurations, the quantity of interest is the \textbf{excessive risk} from pretraining to downstream task: $R_{\text{pre}}(\hat{\phi}) - R^*_{\text{task}}$ where the former is pretraining risk of $\hat{\phi}$.
By definition, the best empirical risk for a pretrained $\hat{\phi}$ is:
\[
R^*_{\text{pre}}:=\min_{f\in\Fcal} \Ebb_{(x,y)\sim P_{\tau}} \ell \big( f\circ\hat{\phi}(x), y \big),
\]
while the optimal downstream risk is given by:
\begin{equation*}
\label{eqn:optimal-risk}
    R^*_{\text{task}} = \min_{\phi \in \Phi} \Ebb_{P_{\tau} \sim \Ecal} \Big[ \min_{f\in\Fcal} \Ebb_{(x,y)\sim P_{\tau}}\ell\big(f\circ\phi(x), y \big) \Big].
\end{equation*}
$R^*_{\text{task}}$ is optimal since it jointly minimizes the representation and downstream hypothesis under the same environment. 


Now we are ready to state the generalization error bound for each configuration. All the proof details in this paper are deferred to the \emph{online material}.

\begin{theorem}
\label{thm:iid-generalization}
We assume $\|\phi(x)\|\leq R$ for all $\phi\in\Phi$. Let $\Rcal_{n}(\Phi)$ and $\Gcal_{n}(\Phi)$ be the empirical Rademacher and Gaussian complexity \cite{bartlett2002rademacher} of $\Phi$ based on the pretraining data.  \\
\textbf{Homogeneous configuration (BERT-LR)}. With probability $1-\delta$ (with respect to the random draws of pretraining sample), it holds:
\begin{equation*}
    R_{\text{task}}(\hat{\phi}) - R^*_{\text{task}} \lesssim \frac{\Gcal_n(\Phi)}{\sqrt{n}} + \frac{R}{n} + \sqrt{\log(8/\delta)}.
\end{equation*} \\
\textbf{Heterogeneous configuration (CL-LR)}. Suppose the optimal pretraining risk is achieved by $\phi^* = \arg\min_{\phi}R^C_{\text{CL}}(\phi)$, it then holds with probability at least $1-\delta$ that:
\begin{equation*}
    R_{\text{task}}(\hat{\phi}) - R^*_{\text{task}} \lesssim \frac{R\Rcal_n(\Phi)}{n} + \frac{R^2\sqrt{\log(1/\delta)}}{\sqrt{n}} + R\cdot\Ebb_{y}\Big\|\text{cov}_{P_X^{(y)}}(\phi^*)\Big\|_2,
\end{equation*}
where $\text{cov}_{P_X^{(y)}}(\cdot)$ denotes the covariance matrix under $P_X^{(y)}$. The exact formulation of this quantity and the complexity terms are deferred to online material.
\end{theorem}

When comparing the two bounds in \ref{thm:iid-generalization}, first observe that \emph{CL-LR} has the additional term of $\Ebb_{y}\big\|\text{cov}_{P_X^{(y)}}(\phi^*)\big\|_2$ which exactly captures the consequence of using different encoding and decoding structures. In particular, this term can become arbitrarily large if $\phi^*$ does not agree well with the conditional distribution of the downstream task, which is likely to occur if the downstream model does not decode $\phi^*$ in the way that it was encoded during pretraining. 

The implication of the second bound is significant: even in situations where pretraining and downstream distributions are identical, using \emph{heterogeneous configuration} may introduce uncertainties from the model side, leading to \emph{a lack of stability and predictability} for downstream performance.

In contrast, the bound for the \emph{homogeneous configuration} does not include any additional term that is caused by structure mismatch. The only non-concentrating term is $\sqrt{\log 8/\delta}$, which is a slack term due to the random nature of finite-data sampling. As a result, the downstream performance is much more stable compared to the \emph{heterogeneous configuration}.

Theorem \ref{thm:iid-generalization} provides a rigorous justification for our intuition regarding how model structure can influence the \emph{stability and predictability} of using pretraining embedding for downstream tasks. Although we only prove the results for two specific configurations, they are sufficient to demonstrate the overlooked impact of \emph{structure mismatch} for downstream generalization.

\textbf{Summary}. Based on our analysis, it is recommended to use the \emph{homogeneous configuration} whenever possible to ensure stability. In practice, the fusion of hidden representations in downstream models can always be configured to achieve an \emph{approximately homogeneous configuration}. Moreover, the analysis presented here suggests a potential avenue for studying \emph{predictability}.
\begin{itemize}[leftmargin=*]
    \item {predictability can benefit from certain \textbf{\emph{alignment}} between the pretraining and downstream model.}
\end{itemize}
In fact, employing the same structure for both pretraining and downstream models is a manifestation of \emph{algorithmic alignment} \cite{sturmfels2008algorithms}. We demonstrate in the following section that certain \emph{alignment quantities} can be established through the kernel perspective of deep learning models.

\section{The kernel view of pretrained embeddings}
\label{sec:kernel}
The inner product of pretrained embeddings can be used to define a kernel function, where $K_{\phi}: \Xcal \times \Xcal \to R$ such that $K_{\phi}(x,x') := \big\langle \phi(x), \phi(x') \big\rangle$. Here, $K_{\phi}$ is positive-semidefinite since it admits a \emph{Gram matrix decomposition} with $\phi(\cdot)$ as the \emph{feature lift} of the kernel \cite{shawe2004kernel}. Note that using $K_{\phi}$ directly in downstream task can be challenging due to the poor stability of kernel methods.
However, the kernel perspective can be extremely helpful in understanding how the pretrained embedding can align with the downstream tasks.


\subsection{Pretrained embedding kernel}
\label{sec:interpretation}

The seminal work by \citet{jacot2018neural} shows that under \emph{gradient descent} (GD), neural networks with large width and scaled normal initialization behave like \textbf{kernelized predictors} during training. We elaborate this idea using entity classification task. Denote by $f\big(\theta; \phi(x)\big)$ the downstream classifier parameterized by $\theta$, which takes the pretrained embedding $\phi(x)$ as input. Under a proper initialization, the \emph{Taylor expansion} for a width-$q$ MLP admits:
\begin{equation*}
\begin{split}
    f\big(\theta;\phi(x)\big)  =  f\big(\theta^{(0)};\phi(x)\big) + \big\langle \theta-\theta_0, \nabla f\big(\theta^{(0)};\phi(x)\big) \big\rangle + \Ocal\big(\sqrt{1/q}\big),
\end{split}
\end{equation*}
where $\theta^{(0)}$ are the values at initialization. Analytically, we can always use reparameterization to make the intercept term vanish, e.g. by finding $f\big(\theta;\,\cdot\big) = g\big(\theta_1;\,\cdot\big) - g\big(\theta_2;\,\cdot\big)$ and letting $\theta_1^{(0)} = \theta_2^{(0)}$. Therefore, as $q$ gets large, we have the following approximation:
\[
f\big(\theta;\phi(x)\big) \approx \big\langle \theta-\theta_0, \nabla f\big(\theta^{(0)};\phi(x)\big) \big\rangle.
\]
Here, $\nabla f\big(\theta^{(0)};\phi(x)\big)$ resembles the role of a feature lift for the \textbf{\emph{neural tangent kernel}} (NTK) defined by: 
\[
\text{NTK}\big(\phi(x), \phi(x')\big) = \big\langle \nabla f\big(\theta^{(0)};\phi(x)\big), \nabla f\big(\theta^{(0)};\phi(x')\big) \big\rangle.
\]
Note that NTK depends entirely on the initialization and model structure, which means it is \textbf{invariant} to the GD optimization. Therefore, the \emph{optimization} and \emph{generalization} properties of $f(\theta; \,\cdot)$ during training is completely characterized by its NTK \cite{bartlett2021deep}. 

In what way is the NTK connected to the embedding kernel $K_{\phi}$? We can consider the example of the \emph{two-layer MLP}, where applying the findings from \citet{cho2009kernel} results in the following relationship:
\begin{equation}
\label{eqn:ntk-mlp}
\text{NTK}\big(\phi(x), \phi(x')\big) = 1 - \frac{1}{\pi}\cos^{-1}\frac{K_{\phi}(x,x')}{\|\phi(x)\|\cdot \|\phi(x')\|}.
\end{equation}
It suggests the NTK is a composition of the \emph{arc-cosine kernel} and the \emph{embedding kernel $K_{\phi}$}. Even in more general cases, \emph{neural tangent kernel} still admits the formulation of $\text{NTK}\big(\phi(x), \phi(x')\big) = g \big(K_{\phi}(x,x') \big)$ for some scalar function $g(\cdot)$. 

We note that this remarkable finding extends to other neural networks with a regular NTK \cite{arora2019exact,yang2019scaling}, including sequential neural networks such as RNN and attention mechanisms \cite{yang2019scaling}. We will not delve into the details here but instead provide a simple example to illustrate this concept.

\begin{example}
\label{example:sequential}
\textup{
Suppose the downstream task is to recover $y\in\Rbb^+$ according the fixed-length sequences of $\vec{s} = (x_1,\ldots,x_k)$. It resembles sequential recommendation where we predict the rating of $x_k$ given the previously interacted sequence. We assume a downstream model which is the linear function of the concatenated pre-trained embeddings, i.e. 
\[f\big(\theta; \vec{s}, \phi\big) = \theta^{\intercal} \big[\phi(x_1), \ldots , \phi(x_k)\big],
\]
We use $\Phibf\in\Rbb^{n\times kd}$ to denote the matrix of concatenated embeddings. Given the outcome $\ybf\in\Rbb^n$, the least-square prediction is:
$\Phibf \big(\Phibf \Phibf^{\intercal} \big)^{-1} \Phibf^{\intercal}\ybf$. Notice that the entries in $\Phibf \Phibf^{\intercal}$ -- which determines the dual solution of the linear model \cite{shawe2004kernel} -- corresponds to a new \textbf{sequence kernel} $K_{\text{seq}}$ composed entirely by $K_{\phi}$:
\begin{equation}
\label{eqn:ntk-seq}
K_{\text{seq}}\big(\vec{s},\vec{s}'\big) = \sum_{i=1}^k K_{\phi}\big(x,x'\big), \text{ where } \vec{s}'=\big(x_1',\ldots,x_k'\big).
\end{equation}
Therefore, the pre-trained embedding kernel $K_{\phi}$ can also impact the performance of the sequential downstream tasks.
}
\end{example}

The NTK theory sheds light on how pretrained embeddings influence the performance of downstream task. 
Essentially, if the embedding kernel $K_{\phi}$ already aligns well with the downstream task, even a simple downstream model can deliver satisfactory performance because its NTK function can be well-behaved. In particular, as $\text{NTK}\big(\phi(x), \phi(x')\big) = g \big(K_{\phi}(x,x') \big)$, finding a good $g(\cdot)$ becomes easier in downstream task.

In the next section, we show that this is indeed the case: \textbf{alignment} between $K_{\phi}$ and downstream task can guarantee the performance of pretrained embedding $\phi$. Moreover, the alignment is invariant to the downstream model structure, making it suitable for creating metrics to access the \emph{predictability} of pretrained embeddings for downstream tasks.

\subsection{Kernel-based metrics for downstream task}
\label{sec:OOD}

We will begin with the downstream entity classification task where the label is $y\in\{-1,+1\}$. Suppose that we have the downstream dataset of $(x, y)$. Consider the simple kernel classifier: 
\[
f_{\phi}(x) = E_{x'}\big[y'k_{\phi}(x,x') \big] \big / \sqrt{\Ebb[k_{\phi}^2]}.
\]
where the expectations are taken with respect to the downstream data distributions. We use $R\big(f_{\phi}\big)$ to denote the downstream risk of $f_{\phi}$ under the $0-1$ loss. By applying a novel argument that relies only on the properties of kernel (the proof is deferred to the online material), we can bound the risk of $f_{\phi}$ as below.

\begin{proposition}
\label{prop:ood-clf}
It holds with probability at least $1-\delta$ that:
\begin{equation}
\label{eqn:OOD-clf}
    R\big(f_{\phi}\big) \leq 1- \frac{\Ebb\big[K_{Y}(y,y') K_{\phi}(x,x') \big] \cdot \sqrt{\delta}}{\sqrt{\Ebb[k_{\phi}^2]}},
\end{equation}
where $K_{Y}(y,y')$ is the downstream task kernel given by $1[y=y']$, and the expectation is taken with respect to the joint distribution of $(x,y)$ and $(x',y')$.
\end{proposition}

Proposition \ref{prop:ood-clf} demonstrates that the effectiveness of a particular $\phi(x)$ in the downstream task depends solely on the degree of \emph{alignment} between the kernel $K_{\phi}$ and the downstream labeling mechanism (represented by $K_Y$). This result holds irrespective of the underlying distributions. When $y$ is not binary, it is a straightforward extension to establish that the bound is still determined by $\Ebb\big[K_{Y}(y,y') K_{\phi}(x,x')\big]$ with some properly defined target kernel $K_Y$.

The key takeaway from Proposition \ref{prop:ood-clf} is that the quantity of $\Ebb\big[K_{Y}(y,y') K_{\phi}(x,x')\big]$ can reasonably predict the downstream performance of $\phi(x)$. It is stable under CL-based pretraining (because it relies only on the inner product between embeddings), and can be computed easily on large datasets. We can use it as a metric to understand the \textbf{predictability} of pretrained embedding for downstream \textbf{entity classification} tasks. 

On the other hand, connecting $K_{\phi}$ to downstream (sequential) recommendation tasks is more involved. The approach we take follows the classical \textbf{likelihood principle} \cite{keener2010theoretical}, which states that given a model (regardless of its correctness), all the information in the data relevant to the model parameters is contained in the likelihood function. 
In other words, we can plug in the pretrained embedding, and use the likelihood function to derive the quantity that characterizes the \emph{alignment}. 
The game plan is as follows:
\begin{enumerate}[leftmargin=*]
    \item propose a general downstream model -- which depend on $\phi$ and can have other model parameters -- that describes the sequential behavior of users;
    \item plug in the pre-trained embeddings $\hat{\phi}$ and use \emph{maximum-likelihood estimation} (MLE) to estimate the remaining parameters.
\end{enumerate}

The first step is to design a general interaction model. We focus particularly on the sequential recommendation tasks due to its importance for e-commerce ML. We point out that the recent literature have identified the \textbf{exposure bias} -- which causes the data to missing-not-at-random (\textbf{MNAR}) -- as a key factor for correctly modelling user feedback \cite{xu2020adversarial,schnabel2016recommendations,yang2018unbiased,liang2016modeling}. Particularly in e-commerce, users are more likely to interact with the items that are exposed to them. As a result, the users' decision making is \emph{shifted} by the exposure distribution, and direct learning from their feedback is subject to bias. 

Following that line of work, we propose a general sequential interaction model that explicit accounts for the \emph{exposure probability} when generating the next interaction:
\begin{equation}
    \label{eqn:interaction}
    p\big(x_{k+1} \,|\, s \big) = \lambda p_0(x_{k+1}) + (1-\lambda) \frac{\exp\big(\big\langle \phi(x_{k+1}), \varphi(s) \big\rangle \big)}{Z_s},
\end{equation}
where $s=(x_1,\ldots,x_k)$ is the sequence of previously interacted items, $p_0$ is the exposure probability, $\varphi(s)$ is the unknown representation of the sequence, and $Z_s = \sum_{x\in\Xcal}\exp\big(\langle \phi(x_{k+1}), \varphi(s)  \rangle \big)$ is the normalizing constant. Here, $\lambda\in (0,1)$ decides to what extend the user is affected by exposure bias. The free parameters of the model is $\varphi(s)$ -- the representation of the past interacted sequence. The model in (\ref{eqn:interaction}) is an exact characterization of how users interacts with the next item according to:
\begin{itemize}
    \item how likely the item is exposed to the user;
    \item how well the item matches user intention (reflected by the sequence of previously interacted items).
\end{itemize}
Following our roadmap, the next step is to obtain the MLE of the sequence embedding $\varphi(s)$ given the pre-trained embeddings $\hat{\phi}$. We are particularly interested in how it relates to $K_{\phi}$, and we reveal it in the following proposition (the proof is in the online material).


\begin{proposition}
\label{prop:MLE}
The maximum-likelihood estimation of $\varphi(s)$ is:
\begin{equation}
    \hat{\varphi}\big(s\big) = \sum_{i=1}^k\frac{\alpha}{p_0(x_i) + \alpha} \phi\big(x_i\big),
\end{equation}
where $\alpha>0$ is a constant that depends on $\lambda$ and $Z_s$.
\end{proposition}

Proposition \ref{prop:MLE} shows that the maximum-likelihood estimation of the sequence representation is a weighted combination of the pretrained embeddings, where the weights reflect the exposure probabilities. This result agrees with the existing arguments from the \emph{data MNAR} literature that the regularly exposed items should be discounted by their exposure probability \cite{schnabel2016recommendations,yang2018unbiased}. More importantly, now the relevance score between the previous interacted sequence and the next candidate item is precisely given by:
\begin{equation}
\label{eqn:seq-score}
    \exp\big(\big\langle \phi(x_{k+1}), \varphi(s) \big\rangle\big) = \sum_{i=1}^k\frac{\alpha}{p_0(x_i) + \alpha} K_{\phi}\big(x_{k+1}, x_i\big),
\end{equation}
which is a simple function of $K_{\phi}$ that characterizes the \textbf{alignment}. Therefore, we can also use (\ref{eqn:seq-score}) to construct metrics for the \textbf{predictability} of pretrained embedding for downstream \textbf{sequential recommendation} tasks. In particular, for ranking problems, we simply use (\ref{eqn:seq-score}) to rank candidate items to obtain the correspond \emph{recall}, \emph{NDCG}, and \emph{MRR} values. Similar to entity classification, the kernel-based metric is \emph{stable} and \emph{easy to compute}. In the next section, we use benchmark datasets and online experiments to examine the predictability of the proposed kernel-based metrics for downstream \emph{entity classification} and \emph{sequential recommendation} tasks.

\begin{figure}[hbt]
    \centering
    \includegraphics[width=\linewidth]{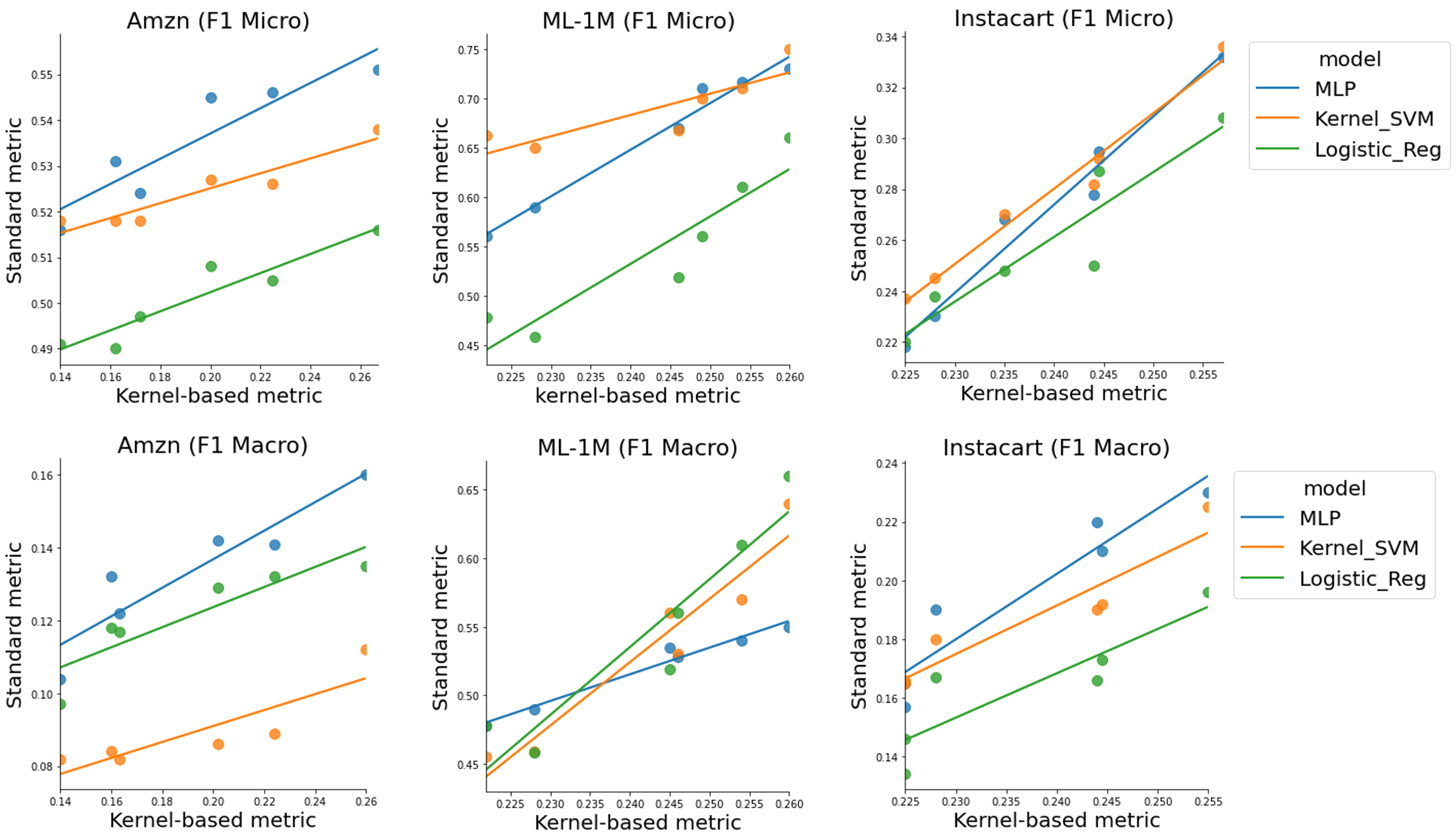}
    \caption{\small The correlation between the kernel-based metric and various models' performances on the content understanding (\emph{entity classification}) task. To visualize the patterns, we generate different sets of pretrained embeddings that lead to the varied performances.}
    \label{fig:OOD-clf}
    \vspace{-0.3cm}
\end{figure}

\subsection{Benchmark experiments and online evaluations of the kernel-based metrics} 

For both downstream tasks, we train the item embeddings in an \emph{Item2vec} fashion \cite{barkan2016item2vec} using the user interaction data. We particularly choose these settings so the two downstream tasks will be OOD in nature considering how the pre-training is conducted. The detailed setups and configurations are deferred to Appendix \ref{append:experiment}. 
In order to obtain the different sets of pretrained embeddings so their kernel-based metrics will cover a wide range of possible performances in the downstream tasks, we vary the \emph{window size} in \{2,3\} and the \emph{\#negative samples} in \{2,3,4\} for the \emph{Item2vec} pretraining. 

In the entity classification task, we use \emph{logistic regression}, \emph{kernel SVM} and \emph{two-layer MLP} as the downstream models. Recall that the kernel-based metric is computed via:
\begin{equation}
\label{eqn:kernel-clf}
\sum_{(x',y')}\Big[K_{Y}(y,y') K_{\phi}(x,x') \Big] \big / \sqrt{\sum_{x,x'}K_{\phi}(x,x')^2},
\end{equation}
where the target kernel is simply given by $K_{Y}(y,y') = 1[y=y']$.

For the downstream sequential recommendation model, we experiment with the MLP4Rec (\textbf{Dense}), GRU4Rec (\textbf{GRU}), and Attn4Rec (\textbf{Attn}) \cite{hidasi2015session,kang2018self}. When computing the kernel-based metric according to (\ref{eqn:seq-score}),we use the items' \emph{log popularity} as a proxy of the exposure probability, which is a common practice for the benchmark datasets we consider here \cite{xu2020adversarial}. We experimented with a wide range of $\alpha$ (the hyperprameter in (\ref{eqn:seq-score})), and find that the outcome is not sensitive to its value since it merely affects the scaling factor. Therefore, we fixed $\alpha$ as the average exposure probability for all of our experiments.

\begin{figure}[hbt]
    \centering
    \includegraphics[width=\linewidth]{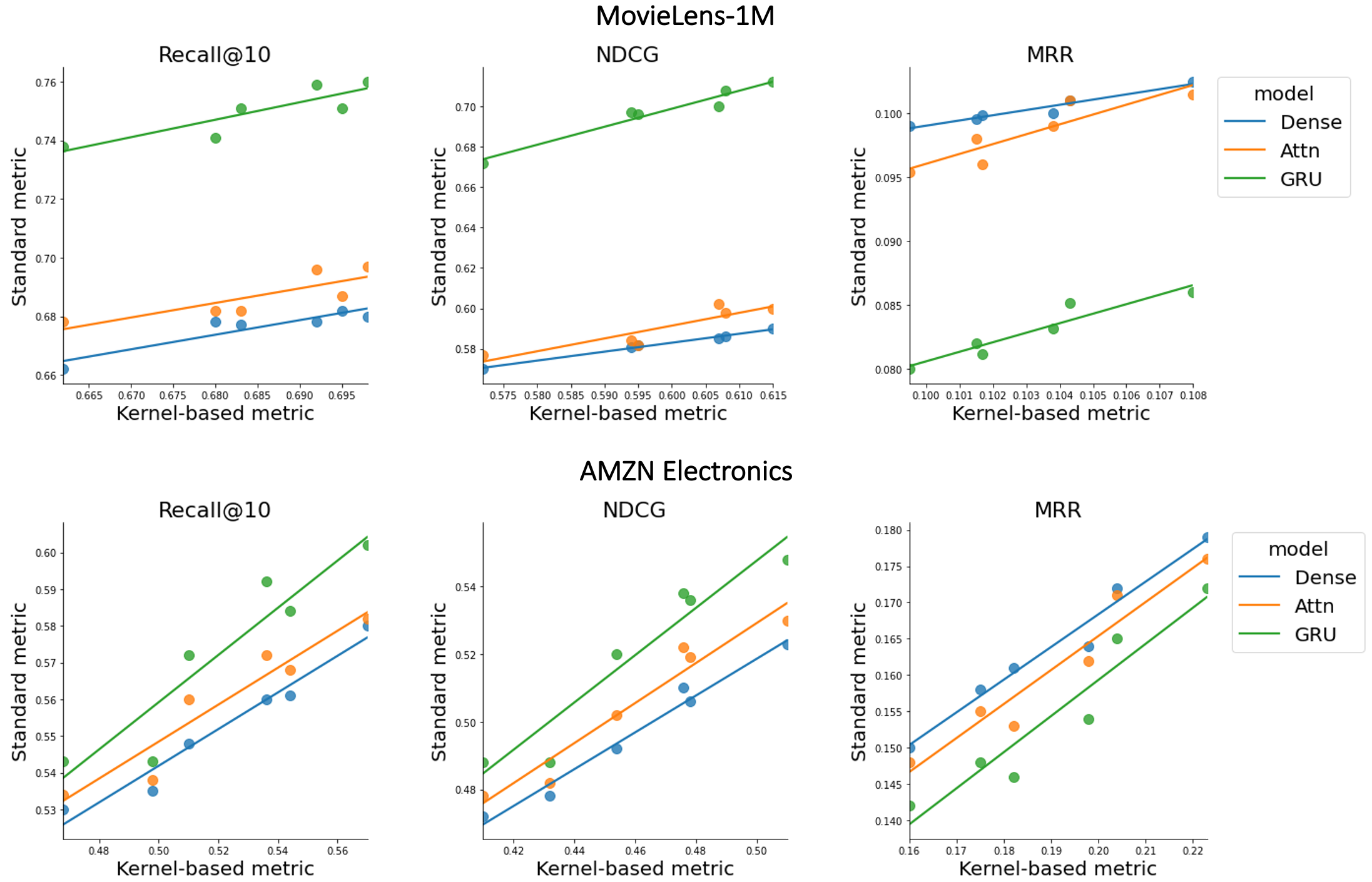}
    \caption{\small The correlation between the kernel-based metric and various models' performances on the \emph{sequential recommendation task} with different sets of pretrained embeddings.}
    \label{fig:OOD-reco}
\end{figure}

The correlation between the kernel-based metrics and the downstream models' performances are visualized in Figure \ref{fig:OOD-clf} and \ref{fig:OOD-reco}. We mention that since the shopping basket of \emph{Instacart} lacks sequential information, we only present results from the other two datasets in the context of sequential recommendation tasks. We observe that for both entity classification and sequential recommendation, there is a \textbf{strong positive correlation} between the kernel-based metrics and the actual downstream performance for all the models and evaluation metrics we considered. These results demonstrate the \textbf{predictive} nature of the proposed metrics in assessing the potential downstream performance of pretrained embeddings.

\textbf{Real-world deployment analysis}.
Since offline evaluations on benchmark datasets are prone to bias, we carry out real-world experimentation with 'ECOM' to examine the effectiveness of our kernel-based metrics for recommendation task. Due to the space limitation, the detailed descriptions and experiment setups are deferred to Appendix \ref{append:online}. 
The context of our deployment is identifying the best-performing pretrained embedding from several candidates for item-page recommendation. The downstream ranking model is sequential and has a \emph{deep \& wide architecture} \cite{cheng2016wide}. 

\begin{figure}[htb]
    \centering
    \includegraphics[width=\linewidth]{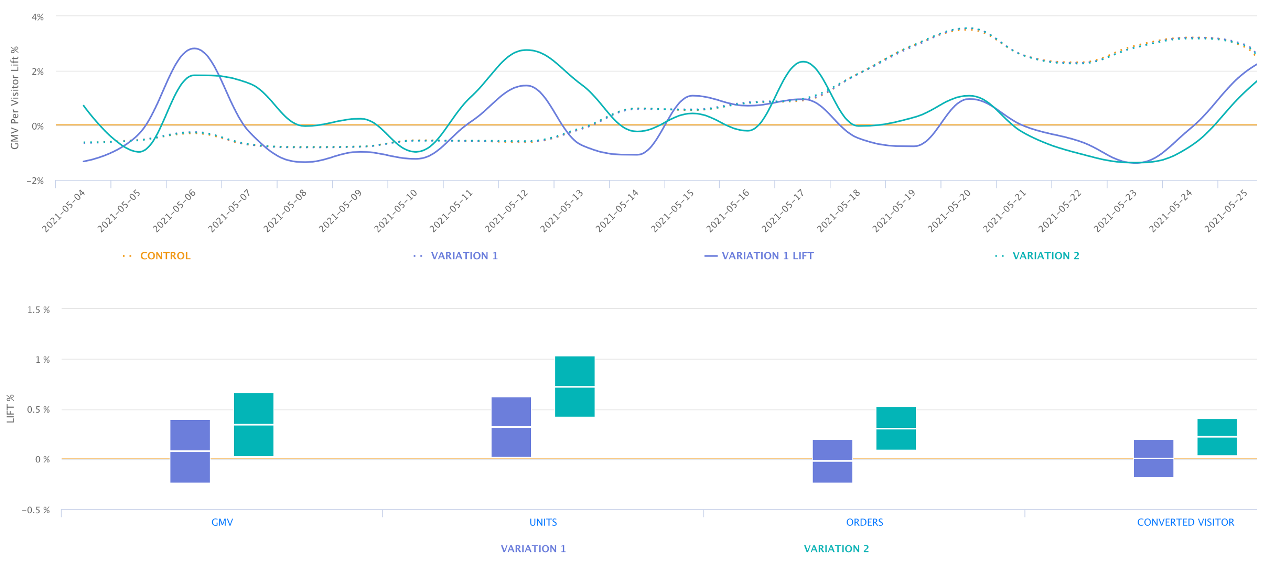}
    \caption{\small Summary of the online A/B/C testing results, for \textbf{Emb1} (\textbf{Control}), \textbf{Emb2} (\textbf{Variation 1}) and \textbf{Emb3} (\textbf{Variation 2}). The {\color{orange}orange} baselines represent the performance of Emb1, the {\color{blue}blue} lines and bars represent the performance of Emb2, the {\color{teal}green} lines and bars represent the performance of Emb3. For all business metrics we monitored (GMV, \#Units, \#Orders, \#Converted Visitors), Emb3 outperformed Emb2 which outperformed Emb1. We mention that the figures are automatically generated by the testing platform so we are unable to change the font sizes.}
    \label{fig:abc-testing}
\end{figure}

This is a common problem in the e-commerce industry, as there are many pretraining solutions available. By understanding their potential downstream performance, the cost of conducting online tests can be reduced. In our case, we have three candidate embeddings:
\begin{itemize}[leftmargin=*]
    \item \textbf{Emb1}: item embedding trained by \emph{Doc2vec} \cite{le2014distributed} using item textual descriptions. This version is currently in production, so it will be the \textbf{control method} in online testing. 
    \item \textbf{Emb2}: complementary item embedding \cite{xu2020knowledge} which better captures the complementary relationship among items.
    \item  \textbf{Emb3}: knowledge-graph-based item embedding \cite{xu2020product} which is enhanced by the relationships contained in the knowledge graph.
\end{itemize}
We launch an A/B/C testing on the 'ECOM' platform to compare the three versions, and the testings results are provided in Figure \ref{fig:abc-testing}. For privacy reasons, we visualize the relative lift of \emph{Emb2} and \emph{Emb3} over the control baseline (\emph{Emb1}). The upper panel of Figure \ref{fig:abc-testing} is the monitoring of the gross merchandise value (GMV) during the testing period, and the lower panel summarizes the final outcome. From the online testing, we conclude that \emph{Emb3} gives the best downstream performance, and \emph{Emb2} slightly outperforms \emph{Emb1}.

\begin{figure}[htb]
    \centering
    \includegraphics[width=1.0\linewidth]{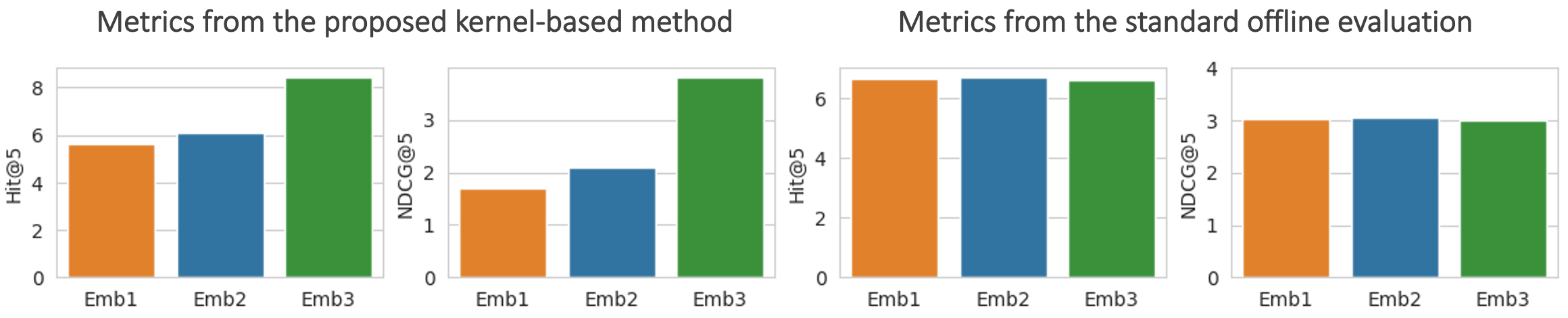}
    \caption{\small The offline evaluations results for comparing the standard offline evaluation metrics and the proposed kernel-based metrics. The reported results have been multiplied by 100. }
    \label{fig:offline-eval}
\end{figure}

In our offline experiments prior to the online testing, we computed the \emph{kernel-based metric} and conducted \emph{standard offline evaluation} with Recall@5 and NDCG@5 as metrics (see Appendix \ref{append:online} for detail). In standard offline evaluation, we must retrain and tune the downstream model for each candidate embedding, which is neither interactive nor scalable. The results are presented in Figure \ref{fig:offline-eval}. The lower panel shows that in standard offline evaluation, the three candidate embeddings perform similarly, indicating that the standard evaluation metrics are not effective predictors of the actual downstream performance of pretrained embeddings. In contrast, the upper panel of Figure \ref{fig:offline-eval} shows that the \emph{kernel-based metrics} can distinguish between the three versions of pretrained embeddings and are highly correlated with the online testing outcomes.


\textbf{Summary}. Our benchmark and online experiments provide additional insight into the ability of kernel-based metrics to predict downstream performance of pretrained embeddings. We found that these metrics provide highly \textbf{interactive} and \textbf{scalable} solutions for selecting pretrained embeddings. Note that traditional offline evaluation requires a separate training and tuning process for each candidate embedding, which is extremely cumbersome. In contrast, the kernel-based metrics are easily computed and can be widely used in real-world applications.




\section{Related Work}
\label{sec:related}

Our work relates to a wide spectrum of recent literature:
\begin{itemize}[leftmargin=*]
    \item \textbf{Understanding algorithmic stability.} While deep learning models are often considered as universal approximators, they carry different algorithmic uncertainties due to their complex structures \cite{soulos2020discovering,bahdanau2018systematic,russin2019compositional,hupkes2020compositionality}, which makes it important to understand how the specific model structure can affect learning stability \cite{akter2021algorithmic,hooker2021moving}. Our work contributes to this venue of research by revealing the impact of algorithmic structures for pretraining, and we provide concrete domain analysis for e-commerce ML. 
    
    \item \textbf{Downstream generalization}. Understanding how the learnt patterns may generalize to downstream data is essential for modern ML applications \cite{hendrycks2021many,krueger2021out,zhou2021domain,shen2021towards}. Our work investigates this problem from a principled kernel perspective, which helps predicting and understanding the capability of pretrained embeddings for downstream tasks. The kernel view we adopted may shed insights to future work in this direction.
    
    \item \textbf{Model explainability}. Compared with other domains such as CV and NLP, explainability in e-commerce ML has received less attention. The few existing literature are associated with causal interpretation \cite{bottou2013counterfactual}, feature-based interpretation \cite{tsang2019feature,lundberg2017unified}, or model-driven interpretations \cite{ribeiro2016should}. While some recent work interpret embeddings under specific domain context \cite{xu2021theoretical,allen2019analogies}, we focus on the general interplay between pretrained embeddings and downstream tasks and the impact of pretraining-downstream configurations. The kernel interpretation is also useful in its own regard.
    
    \item \textbf{Exposure bias in recommendation}. Dealing with exposure bias is critical for improving the learning and evaluation performance of e-commerce recommendation \cite{schnabel2016recommendations,yang2018unbiased,liang2016modeling,xu2020adversarial}. Unlike the existing solution that rely on implicit weighting, we explicitly account for the exposure probability in the data generating process. Our result derived from maximum likelihood shows the impact of an individual entity is discounted by its exposure probability, which provides another angle for building recommendation models that handle exposure bias.
    
\end{itemize}

\section{Discussion}

We address several key challenges that arise when utilizing pretrained embeddings in e-commerce machine learning systems, such as stability and predictability. The conclusions we draw from our study are backed by both theoretical results and experimental evidence (both offline and online). The significance of our work extends beyond the realm of e-commerce machine learning, and can inform future research and practices in building and maintaining robust machine learning systems that make use of pretrained embeddings. While our work may not comprehensively address all practical considerations, our aim is to raise awareness and provide valuable insights for further exploration in leveraging deep learning advancements for industrial applications.

\balance
\bibliographystyle{ACM-Reference-Format}
\balance
\bibliography{references}

\appendix
\setcounter{equation}{0}
\renewcommand{\theequation}{A.\arabic{equation}}
\setcounter{figure}{0}
\renewcommand{\thefigure}{A.\arabic{figure}}
\setcounter{table}{0}
\renewcommand{\thetable}{A.\arabic{table}}
\setcounter{claim}{0}
\renewcommand{\theclaim}{A.\arabic{claim}}
\setcounter{proposition}{0}
\renewcommand{\theproposition}{A.\arabic{proposition}}
\setcounter{lemma}{0}
\renewcommand{\thelemma}{A.\arabic{lemma}}
\setcounter{theorem}{0}
\renewcommand{\thetheorem}{A.\arabic{theorem}}

\newpage
\section{Details for the benchmark experiments}
\label{append:experiment}

We present the experiment details in this part of the paper. When unspecificed, the embedding dimension is given by $d=32$.

\subsection{Model Configurations}

\begin{figure}[bht]
    \centering
    \includegraphics[width=0.6\linewidth]{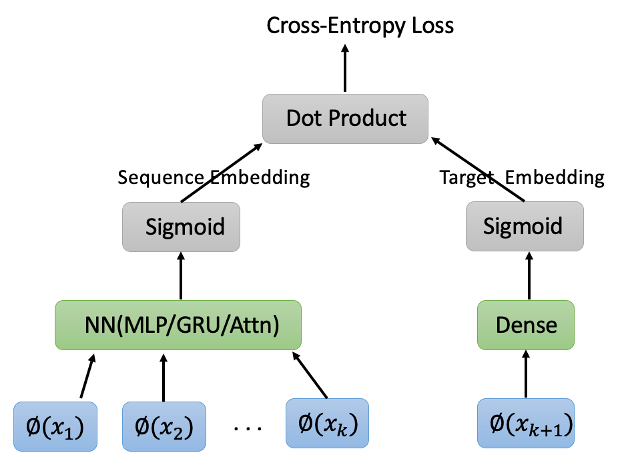}    
    \caption{\small The two-tower architecture we employed for the downstream sequential recommendation task on benchmark datasets. As we discussed in Section \ref{sec:structure}, we should use \emph{homogeneous configuration} whenever possible. In the two-tower architecture, the \emph{fusion} of the two towers is where we can (approximately) achieve \emph{homogeneous configuration}. Note that we use the word 'approximately' because there are model components (e.g. RNN) we do not wish to change. Since the pretraining model uses inner product between embeddings, here we also use dot product for the final fusion layer.}
    \label{fig:two-tower}
\end{figure}

In our experiments, we mainly employ the \emph{Item2vec} model \cite{barkan2016item2vec}, the \emph{BERT} model \cite{devlin2018bert}, and the two-tower \emph{Dense-layer-based}, \emph{GRU-based}, and \emph{attention-based} recommendation models to corroborate our analysis.
\begin{itemize}[leftmargin=*]
    \item \textbf{Item2vec}: the model is a direct adaptation of the renown NLP Word2vec model \cite{mikolov2013distributed}, by replacing the original word sequence by user behavior sequence. Therefore, the algorithm takes the \emph{window size} and \emph{\#negative samples} as hyper-parameters, which we vary between $\{2,3\}$ and $\{2,3,4\}$, respectively, to generate pretrained embeddings whose performances will cover a wide spectrum of the metrics we consider (as we showed in Figure \ref{fig:OOD-clf} and \ref{fig:OOD-reco}).
    
    \item \textbf{BERT}: since the sentence and descriptions for the items are relatively short, we use the \emph{ALBERT} (a simplified version of BERT) whose pretrained version is publicly available, together with its dedicated NLP preprocessing pipeline that is ready for retraining the final dense layer \footnote{\url{https://www.tensorflow.org/official_models/fine_tuning_bert}}. Note that retraining the final dense layer is equivalent to using the second-to-last hidden layer as the embedding and apply a linear (logistic) regression on top of it. In our paper, we rephrase this \emph{linear-head-tuning} approach as \textbf{BERT-LR}.
    
    \item \textbf{Two-tower recommendation models (e.g. Dense4Rec, GRU4Rec, Attn4Rec)}: the two-tower architecture is a popular model choice for sequential recommendation (visualized in Figure \ref{fig:two-tower}). In particular, we use the \emph{Dense layer}, \emph{GRU layer}, and \emph{self-attention layer} to aggregate the users' previous interaction sequences. Since our experiments primarily serve illustration purposes, we directly fix the hidden dimensions of all the Dense layers as \{32,16\}, and the hidden dimension of the GRU and self-attention layer as 32. The candidate item embedding goes through a dense layer, and the fusion of the two towers is done by taking the inner product between their hidden representations.
\end{itemize}

Recall that the pretraining-downstream model configurations we used include: \textbf{CL-LR}, \textbf{CL-IP}, \textbf{BERT-LR}, \textbf{BERT-IP}. In particular, \textbf{CL-LR} means we use the CL-pretrained embeddings as features for the downstream logistic regression model, \textbf{CL-IP} means we use the inner product of pretrained embeddings to compute the kernel $K_{\phi}$,  and use $K_{\phi}$ for a \emph{kernel SVM model}. As for \textbf{BERT-IP}, we simply extract the second-to-last layer as pretrained embedding, and proceed in the same way as \emph{CL-IP} where we first build the pretrained embedding kernel $K_{\phi}$ and then employ the kernel SVM.

\subsection{Training, Validation, Evaluation and Computation}

We describe the training, validation, evaluation, and computation details of our benchmark experiments. We start with the illustration experiment that we conducted for Figure \ref{fig:emb-var}.

\textbf{Experiment detail for Figure \ref{fig:emb-var}}. For each dataset (ML-1m, Instacart, Amazon), we use the Item2vec model to pretrain item embeddings, using: \emph{window size=3} and \emph{\#negative sample=3}. We use the \emph{Glorot} embedding initialization from \emph{Tensorflow}, and use the \emph{RMSprop} optimizer with initial learning rate as 0.005. The batch size is 256, and the $\ell_2$ regularization is set to 1e-6. We also use early stopping which halts the training when the training accuracy stops improving more than 1e-6 for three consecutive epochs. We repeat the same training process independently for ten time, then randomly pick one item (movie) embedding from each dataset and visualize its values together with the standard deviation error bar.

In what follows, we provide the implementations details for the rest of the benchmark experiments in our paper.

\textbf{Training.} We use off-the-shelf \emph{Tensorflow} implementation for all the deep learning models (including \emph{Item2vec}) mentioned in our paper. The code are available in the online material. We employ the base version of the pre-trained \emph{ALBERT} model \footnote{\url{https://github.com/google-research/albert}} and its official NLP preprocessing pipeline to obtain the BERT-based embeddings using the description (title, context) metadata of each dataset. The \emph{inner-product kernel SVM} (IP) model and \emph{logistic regression} (LR) are implemented using the Scikit-Learn \footnote{\url{https://scikit-learn.org/stable/}} package. We use the stochastic gradient descent (\emph{RMSprop} optimizer) for all the deep learning models, with the learning rate set as 0.005, batch size set as 256, and \emph{Glorot} initilizations when needed. We also use the $\ell_2$ regularization on all the model parameters with the regularization parameter set as $1e-6$ with no decaying schedule.

\textbf{Validation}. For the item classification tasks, we perform validation on the 10\% left-out samples using the \emph{F1 score} as metric. For the sequential recommendation tasks, for each sequence, we use the last interaction for testing, the second-to-last for validation, and the rest for training. We use \emph{Recall@10} as the validation metric for recommendation tasks.

\textbf{Evaluation}. For the item classification tasks, the evaluation metrics \emph{Micro-F1} and \emph{Macro-F1} are computed on the 10\% testing samples also using the \emph{Scikit-Learn} package. The recommendation algorithms are evaluated on the last interaction, where we rank the true interacted among all the items, and compute the top-10 \emph{Recall} (\emph{Hit@10}), overall \emph{NDCG} and the mean reciprocal rank \emph{MRR}. We mention that \citet{krichene2020sampled} has pointed out the potential issue of using sampled metric, so when computing the recommendation metrics, we rank \emph{all} the candidate items to obtain a less biased result.

\textbf{Computation}. Using the kernel-based approach to compute the metrics is straightforward. Given the pretrained embeddings and data from downstream task, we compute the classification score according to (\ref{eqn:kernel-clf}) or rank the candidate items using (\ref{eqn:seq-score}), and then compute the evaluation metrics in the standard way. We can also use random downsampling to further improve the computation efficiency.
All the implementation are carried out in \emph{Python} and we use \emph{Tensorflow} to train the deep learning models. The computations are conducted on a Linux cluster with 16 CPU threads and 128 Gb memory. We also use two Nvidia Tesla V100 GPU.

\section{Online Experiment at 'ECOM'}
\label{append:online}

\begin{remark}
To avoid potential privacy and legal issues with disclosing business information, here we provide necessary background and context about our production problem so the readers can grasp the key challenges we face as well as the value of our solution.
\end{remark}

\textbf{Background}: in a few sentences, the production request is to provide a \emph{centralized pretrained embedding service} to support multiple downstream applications on 'ECOM', and \emph{item-page recommendation} is one of the most important use cases. As a service, we allow downstream model owner to specify the pretraining model and data, using specific rules and languages we developed off-the-shelf. After our service executes the pretraining process, the pretrained embeddings will be passed to the backend \emph{feature store} for offline model development, as well as the \emph{frontend memcache} for online inference. Therefore, \emph{stability} and \emph{predictability} of pretrained embedding are crucial to the whole ML ecosystem.

\textbf{Item-page recommendation system at 'ECOM'}: an item page at 'ECOM' is where customers can view the detailed information an a particular \emph{anchor item} they are interested in. Item-page recommendation aims to provide recommendations that are \emph{relevant} to both the \emph{anchor item} and the customer's \emph{hidden intention}. In general, we consider the customer's previous viewing sequences and some other online features as containing information about \emph{hidden intention}, and we use them to \emph{personalize} the list of relevant items we show to the customer. The recommendation system thus consists of two stages:
\begin{itemize}[leftmargin=*]
    \item \textbf{recall stage} that retrieves a pool of candidate items that are relevant to the anchor item;
    
    \item \textbf{ranking stage} that uses the customer's previous viewing sequences and some other online features to re-rank the pool of candidate items.
\end{itemize}
The recall-rerank solutions are very common in the industry so we do not further introduce them here. In our context, the pretrained item embeddings is primarily used by the reranking algorithm, so its stability and predictability often determines the success of the whole system.

\textbf{Offline evaluations}: for standard offline evaluation, the downstream model owner need to retrain the entire model given a new version of pretrained embeddings. Then they conduct a '\emph{replay}' method to evaluate the new reranking algorithm, where techniques such as \emph{inverse propensity weighting} are used to (approximately) ensure the unbiasedness of the evaluation outcome. We mention that each retraining uses the same exact data, model, and training configurations so the only difference is the version of the pretrained embedding. For kernel-based evaluation, we simply compute the metrics by using (\ref{eqn:seq-score}) to rank the candidate items. We observe that it has a huge computation advantage compared with the standard offline evaluation, and it is able to isolate the impact of pretrained embeddings from other confounding factors such as additional features and the downstream model structure. As we discussed for Figure \ref{fig:offline-eval}, in our example, standard offline evaluation can barely differentiate the three versions, while the kernel-based metrics can correctly predict their relative performances.

\textbf{Online testings}: in our online A/B/C testing, eligible customers are pre-allocated randomly to different treatment and control methods, and we gradually ramp up the traffics that are assigned to the treatment methods. When computing the metrics associated with the online testing, unqualified samples from such as bot behaviors are eliminated. The online testing lasted three weeks, and we observed that one of the treatment methods significantly outperforms the control method under a significance level of 0.05 (shown in Figure \ref{fig:abc-testing}). The online testing results show that the kernel-based metrics are able to predict and differentiate the actual performances of the pretrained embeddings.


\end{document}


\setcounter{equation}{0}
\renewcommand{\theequation}{A.\arabic{equation}}
\setcounter{figure}{0}
\renewcommand{\thefigure}{A.\arabic{figure}}
\setcounter{table}{0}
\renewcommand{\thetable}{A.\arabic{table}}
\setcounter{claim}{0}
\renewcommand{\theclaim}{A.\arabic{claim}}
\setcounter{proposition}{0}
\renewcommand{\theproposition}{A.\arabic{proposition}}
\setcounter{lemma}{0}
\renewcommand{\thelemma}{A.\arabic{lemma}}
\setcounter{theorem}{0}
\renewcommand{\thetheorem}{A.\arabic{theorem}}

\maketitle

\appendix

We present the experiment details in this part of the paper. When unspecificed, the embedding dimension is given by $d=32$.

\section{Model configurations}

In our experiments, we main employ the \emph{Item2vec} model \cite{barkan2016item2vec}, the \emph{BERT} model \cite{devlin2018bert}, and the two-tower \emph{Dense-layer-based}, \emph{GRU-based}, and \emph{self-attention-based recommendation models}.
\begin{itemize}[leftmargin=*]
    \item \textbf{Item2vec}: the model is a direct adaptation of the renown NLP Word2vec model \cite{mikolov2013distributed}, by replacing the original word sequence by user behavior sequence. Therefore, the algorithm takes the \emph{window size} and \emph{\#negative samples} as hyper-parameters, which we vary between $\{2,3\}$ and $\{2,3,4\}$, respectively, to generate pre-trained embeddings whose performances will cover a wide spectrum of the metrics we consider (showed in Figure \ref{fig:OOD-clf} and \ref{fig:OOD-reco}).
    \item \textbf{BERT}: since the sentence and descriptions for the items are relatively short, we use the \emph{ALBERT} model (a simplified version of BERT) whose pre-trained version is publicly available, together with the dedicated NLP preprocessing pipeline for fine-tuning \footnote{\url{https://www.tensorflow.org/official_models/fine_tuning_bert}}. During the fine-tuning, we only re-train the last layer of ALBERT, which is equivalent to applying a logistic regression using second-to-last hidden layer's output as input features We refer to this \emph{linear-head-tuning} setting as \textbf{BERT-LR}.
    \item \textbf{Two-tower recommendation models}: the architecture of the model is presented in Figure \ref{fig:online_eval}a, where we use the \emph{Dense layer}, \emph{GRU layer}, and \emph{self-attention layer} to aggregate the users' previous interaction sequence. Since our experiments are mostly illustrational, we fix the hidden dimensions of all the Dense layers as \{32,16\}, the hidden dimension of the GRU and self-attention layer as 32, instead of tuning them exhaustively. 
\end{itemize}

The model variations that we created include: \textbf{CL-LR}, \textbf{CL-IP}, \textbf{BERT-LR}, \textbf{BERT-IP}. In particular, \textbf{CL-LR} means we use the CL-trained embeddings as features for the downstream logistic regression model, \textbf{CL-IP} means we use the pre-trained embeddings to compute the kernel $K_{\phi}$,  and use $K_{\phi}$ for a \emph{standard kernel SVM model}. As for \textbf{BERT-IP}, we simply extract the second-to-last layer's hidden representation as pre-trained embedding, and follow the same procedure as \textbf{CL-IP}. 

We mention that for the illustrational experiments in Section 3, we use $d=16$ for the sake of visualization. For each dataset, we train the Item2vec model as described above, using: window size=3 and \#negative sample=3. We then randomly pick one item embedding from each dataset, and show its values from the ten independent runs.

\subsection{Training, validation and evaluation}

We describe the training, validation, and evaluation procedures for our experiments.

\textbf{Training.} We use off-the-shelf Tensorflow implementation for the \emph{Item2vec} model and the sequential recommendation models. The code are available in the online material\footref{footnote1}. We use the base version of the pre-trained \emph{ALBERT} model \footnote{\url{https://github.com/google-research/albert}} and its official NLP preprocessing pipeline to obtain the BERT-based embeddings, which are also implemented in Tensorflow. For the CL-based pre-training for entity classification, we use the \emph{Gensim} python library\footnote{\url{https://radimrehurek.com/gensim/parsing/preprocessing.html}} for preprocessing the NLP data. The \emph{kernel SVM }model and \emph{logistic regression} for the above-mentioned \textbf{CL-X} downstream models are implemented using the Scikit-Learn \footnote{\url{https://scikit-learn.org/stable/}} package. We use the stochastic gradient descent (\emph{RMSprop} optimizer) for all the deep learning models, with the learning rate set as 0.005, batch size set as 256,  and \emph{Glorot} initilizations when needed. We also use the $\ell_2$ regularization on all the model parameters with the regularization parameter set as $1e-6$ with no decaying schedule.

\textbf{Validation}. For the item classification tasks, we perform validation on the 10\% left-out samples using the F1 score as metric. For the sequential recommendation tasks, for each sequence, we use the last interaction for testing, the second-to-last for validation, and the rest for training. We use the Recall@10 metric for validation as well.

\textbf{Evaluation}. For the item classification tasks, the evaluation metrics \emph{Micro-F1} and \emph{Macro-F1} are computed on the 10\% testing samples using the \emph{Scikit-Learn} package. The recommendation algorithms are evaluated on the last interaction, where we rank the true interacted among all the items, and compute the top-10 \emph{Recall}, overall \emph{NDCG} and the mean reciprocal rank \emph{MRR}. We mention that \citet{krichene2020sampled} has pointed out the potential issue of using sampled metric, so when computing the NDCG, we include all the candidate items to obtain unbiased result.

All the implementation are carried out in \emph{Python}, and the computations are conducted on a Linux machine with 16 CPU, 64 Gb memory and two 32Gb Nvidia Tesla V100 GPU.

    

\begin{figure}[htb]
    \centering
    \includegraphics[width=0.5\linewidth]{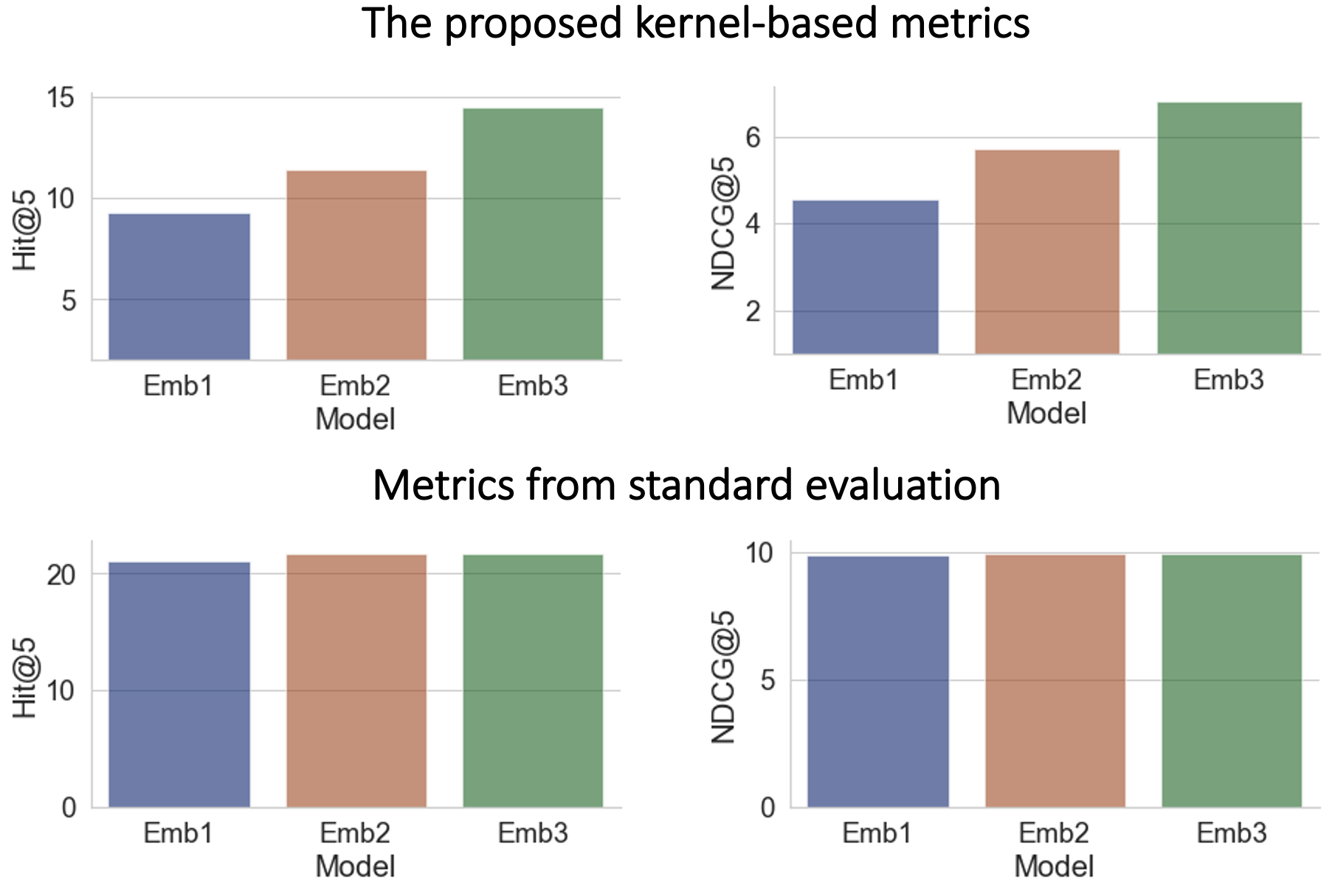}
    \caption{The offline evaluations results for comparing the standard offline evaluation (after correcting for the popularity bias) and the proposed kernel-based evaluation. The reported results have been multiplied by 100. While the standard offline evaluation is unable to differentiate the performance of the three candidate pre-trained embeddings, our proposed kernel-based evaluation clearly shows that \textbf{Emb3>Emb2>Emb1}, which matches the result we observed after deploying online (Figure \ref{fig:abc-testing}).}
    \label{fig:offline-eval}
\end{figure}


\begin{figure}[htb]
    \centering
    \includegraphics[width=0.5\linewidth]{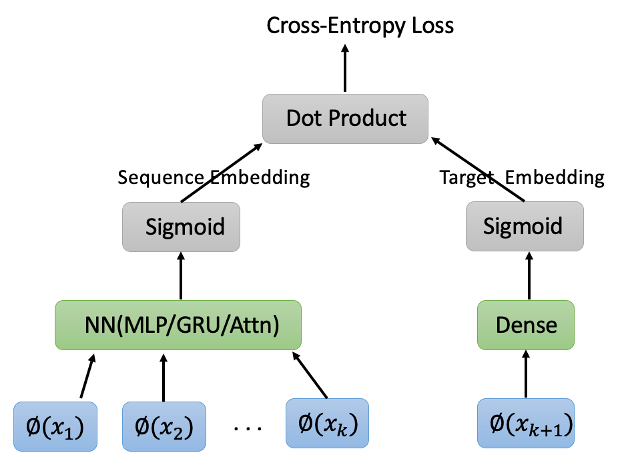}
    \caption{The two-tower architecture for our sequential recommendation models in this paper. }
    \label{fig:two-tower}
\end{figure}

\section{Real-world experiment and Online testing result}
\label{append:online}

The previous results have provided strong evidence to carry the kernel-based evaluation further to online experiments, which we conduct with \emph{'ECOM'} -- a major e-commerce platform in the U.S. The production scenario is that on the item page of \emph{'ECOM'}, we \emph{rank the recall set} according to the customers' most recent ten views. The recall set are fixed during the online experiments, so the role of pre-trained embeddings are strictly for ranking.

The deployed online ranking model also adopts a similar two-tower architecture described in Figure \ref{fig:two-tower}. The difference is that it also processes the non-embedding features such as rating, popularity, and price range, in a similar fashion described in the \emph{Deep \& Wide model} \cite{cheng2016wide}. 
The production problem is to select the best-performing pre-trained item embeddings obtained from the three candidate pre-training algorithms: 
\begin{itemize}[leftmargin=*]
    \item \textbf{Emb1}: the item embedding trained by \emph{Doc2vec} \cite{le2014distributed} that uses only the items' title, brand, and other textual description. This pre-training model is currently in production, so it will serve as the \textbf{control model} in our online experiments. 
    \item \textbf{Emb2}: the complementary item embedding described in the recent work of \cite{xu2020knowledge}, which better captures the complementary relationship among items. This approach also includes the contextual features, so it is a generalization of the previous model
    \item  \textbf{Emb3}: the knowledge-graph-based item embedding that is described in \cite{xu2020product}. In general, this approach enhances traditional item embeddings using techniques that generate knowledge graph embeddings. 
\end{itemize}

\begin{figure}[htb]
    \centering
    \includegraphics[width=\linewidth]{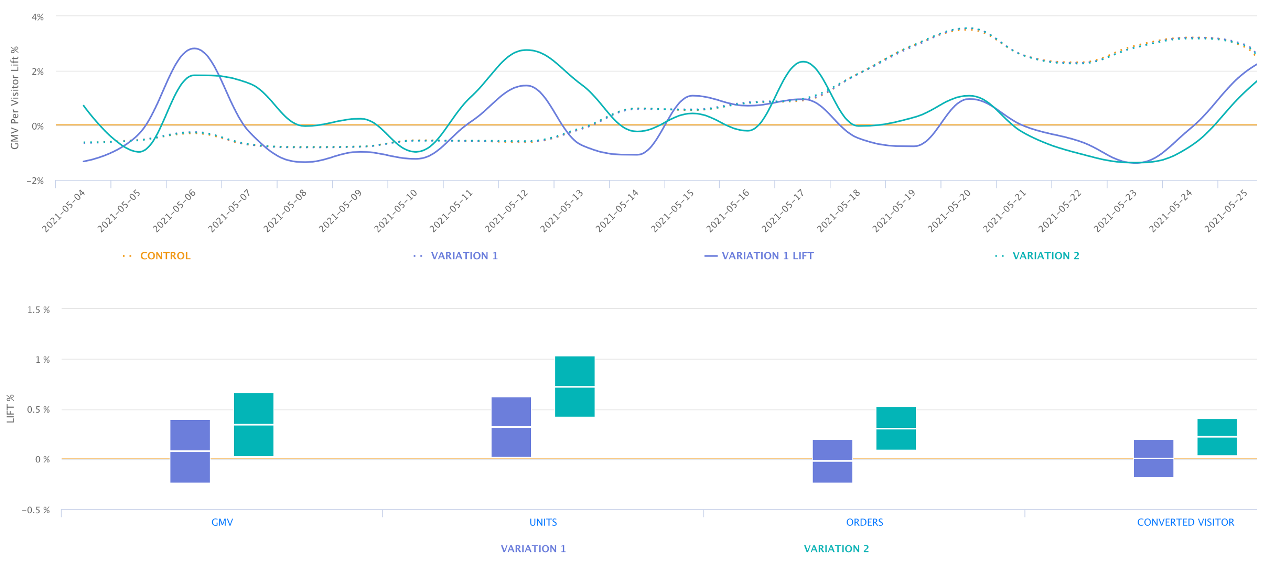}
    \caption{The online A/B/C testing result with \textbf{Emb1} (\textbf{Control}), \textbf{Emb2} (\textbf{Variation 1}) and \textbf{Emb3} (\textbf{Variation 2}), in the dowsntream item-page classification task deployed to 'ECOM'. The orange baselines represent the performance of Emb1, the blue line and bar represent the performance of Emb2, and the green line and bar represent the performance of Emb3.}
    \label{fig:abc-testing}
\end{figure}

To carry out the standard offline evaluation, we first plug in each candidate embeddings and \textbf{retrain} the whole downstream Deep\&Wide model. Retraining the downstream model is necessary as we discussed earlier. We then compute the offline metrics after correcting for the popularity bias \cite{schnabel2016recommendations}, i.e. using popularity to inversely weighting individual samples. 

It is obvious that this standard procedure is very inefficient for evaluating pre-trained embeddings -- we must retrain and tune each model separately to achieve an overall fair comparison. \textbf{More importantly}, due to the existence of other features, it is hard to justify only by the offline result whether the differences owe entirely to the embeddings' quality. 

In this regard, the proposed kernel-based evaluation can connect the performance more straightforwardly to the pre-trained embeddings, and it enjoys faster computation. The only potential concern is that the kernel-based evaluation may not truly imply the online performance. To resolve this concern, we conduct both offline and online evaluations, and compare the offline evaluation results with the online A/B/C testing outcome. They are provided in Figure \ref{tab:offline_eval} and Figure \ref{fig:abc-testing}.

Judging from the kernel-based evaluation in Figure \ref{tab:offline_eval}, we clearly have: \textbf{Emb3}>\textbf{Emb2}>\textbf{Emb1} in terms of both the recall and NDCG (Table \ref{tab:offline_eval}). On the other hand, the comparisons are almost edge-to-edge in the standard offline evaluation. The three pre-trained embeddings perform almost the same, and the margins are very thin.
While we did not expect the kernel-based evaluation to be so advantageous in telling the candidates apart, this result imply another pro of the kernel-based metric: 
\begin{itemize}
    \item they attribute the difference in performances exclusively to the quality of pre-trained embeddings.
\end{itemize}
Finally, from the online A/B/C testing result presented in Figure \ref{fig:abc-testing}, we indeed observe from real-world performances that \textbf{Emb3}>\textbf{Emb2}>\textbf{Emb1} in terms of all the revenue-critical metrics -- \emph{GMV}, \emph{number of orders}, and \emph{conversion rate}. The deployment result shows that the kernel-based evaluation provides efficient and reliable examinations in terms of how the pre-trained embeddings fits the downstream tasks. Nevertheless, we do point out that the kernel-based evaluation is not meant to replace other offline evaluations. Instead, it adds to the arsenal a convenient and powerful tool for the practitioners to examine and understand pre-trained embeddings.

\section{Proofs}

We provide the proofs for the propositions and theorem stated in the paper. 

\subsection{Proof for Proposition 1}

\begin{proof}
The proofs for the lower bound often starts by converting the problem to a hypothesis testing task. Denote our parameter space by $\Bcal(k) = \{\beta \in \Rbb^d: \|\beta\|_0 \leq k\}$. The intuition is that suppose the data is generated by: (1). drawing $\beta$ according to an uniform distribution on the parameter space; (2). conditioned on the particular $\beta$, the observed data is drawn. Then the problem is converted to determining according to the data if we can recover the underlying $\beta$ as a canonical hypothesis testing problem. 

For any $\delta$-packing $\{\beta_1,\ldots,\beta_M\}$ of $\Bcal(k)$, suppose $B$ is sampled uniformly from the $\delta$-packing, then following a standard argument of the Fano method \cite{wainwright2019high}, it holds that:
\begin{equation}
\label{eqn:prop-1}
    P\big(\min_{\hat{\beta}}\sup_{\|\beta^*\|_0 \leq k}) \|\hat{\beta} - \beta^*\|_2 \geq \delta/2 \big) \geq \min_{\tilde{\beta}} P\big(\tilde{\beta} \neq B \big),
\end{equation}
where $\tilde{\beta}$ is a testing function that decides according to the data if the some estimated $\beta$ equals to an element sampled from the $\delta$-packing. The next step is to bound $\min_{\tilde{\beta}} P\big(\tilde{\beta} \neq B \big)$, whereas by the information-theoretical lower bound (Fano's Lemma), we have:
\begin{equation}
\label{eqn:prop-2}
    \min_{\tilde{\beta}} P\big(\tilde{\beta} \neq B \big) \geq 1-\frac{I(y,B) + \log 2}{\log M},
\end{equation}
where $I(\cdot, \cdot)$ denotes the mutual information. Then we only need to bound the mutual information term. Let $P_\beta$ be the distribution of $\ybf$ (which the vector consisting of the $n$ samples) given $B=\beta$. Since $\ybf$ is distributed according to the mixture of: $\frac{1}{M}\sum_i P_{\beta_i}$, it holds:
\[
I(y, B) = \frac{1}{M}\sum_{i}D_{KL}\big(P_{\beta_i} \| \frac{1}{M}\sum_{j}P_{\beta_j}\big) \leq \frac{1}{M^2} \sum_{i,j} D_{KL}\big(P_{\beta_i} \| P_{\beta_j}\big),
\]
where $D_{KL}$ is the Kullback-Leibler divergence.
The next step is to determine $M$: the size of the $\delta-$packing, and the upper bound on $D_{KL}\big(P_{\beta_i} \| P_{\beta_j}\big)$ where $P_{\beta_i}, P_{\beta_j}$ are elements of the $\delta-$packing. 

For the first part, it has been shown that there exists a $1/2$-packing of $\Bcal(k)$ in $\ell_2$-norm with $\log M\geq \frac{k}{2}\log\frac{d-k}{k/2}$ \cite{raskutti2011minimax}. As for the bound on the KL-divergence term, note that given $\beta$, $P_{\beta}$ is a product distribution of the condition Gaussian: $y|\epsilon \sim N\big(\beta^{\intercal}\epsilon \frac{\sigma_{z}^2}{\sigma_{\phi}^2}, \beta^{\intercal}\beta(\sigma_z^2 - \sigma_z^4/\sigma_{\phi}^2)\big)$, where $\sigma_{\phi}^2 := \sigma_{z}^2 + \sigma_{\epsilon}^2 $. 

Henceforth, for any $\beta_1, \beta_2 \in \Bcal(k)$, it is easy to compute that:
\begin{equation*}
\begin{split}
    & D_{KL}(P_{\beta_1}\|P_{\beta_2}) \\
    & = \Ebb_{P_{\beta_1}}\Big[ \frac{n}{2}\log\Big(\frac{\beta_1^{\intercal}\beta_1(\sigma_z^2 - \sigma_z^4/\sigma_{\phi}^2)}{\beta_2^{\intercal}\beta_2(\sigma_z^2 - \sigma_z^4/\sigma_{\phi}^2)} \Big) + \frac{\big\|\ybf-\beta_2^{\intercal}\epsilonbf \frac{\sigma_{z}^2}{\sigma_{\phi}^2}\big\|_2^2}{2\beta_2^{\intercal}\beta_2(\sigma_z^2 - \sigma_z^4/\sigma_{\phi}^2)} - \frac{\big\|\ybf-\beta_1^{\intercal}\epsilonbf \frac{\sigma_{z}^2}{\sigma_{\phi}^2}\big\|_2^2}{2\beta_1^{\intercal}\beta_1(\sigma_z^2 - \sigma_z^4/\sigma_{\phi}^2)} \Big] \\
    & = \frac{\sigma_z^2}{2\sigma_{\epsilon}^2}\|\epsilonbf(\beta_1 - \beta_2)\|_2^2,
\end{split}
\end{equation*}
where $\ybf$ and $\epsilonbf$ are the vector and matrix consists of the $n$ samples, i.e. $\ybf\in\Rbb^n$ and $\epsilonbf\in\Rbb^{n\times d}$, 
Since each row in the matrix $\epsilonbf$ is drawn from $N(0, \sigma_{\epsilon}^2I_{d\times d})$, standard concentration result shows that with high probability, $\|\epsilonbf(\beta_1 - \beta_2)\|_2^2$ can be bounded by $C\|\beta_1 - \beta_2\|_2^2$ for some constant $C$. It gives us the final upper bound on the KL divergence term:
\[
D_{KL}(P_{\beta_1}\|P_{\beta_2}) \lesssim \frac{n\sigma_z^2\delta^2}{2\sigma_{\epsilon}^2}.
\]

Substitute this result into (\ref{eqn:prop-2}) and (\ref{eqn:prop-1}), by choosing $\delta^2 = \frac{Ck\sigma_{\epsilon}^2}{\sigma_z^2n}\log\frac{d-k}{k/2}$ and rearranging terms, we obtain the desired result that with probability at least $1/2$:
\[
\inf_{\hat{\beta}} \sup_{\beta^*: \|\beta^*\|_0 \leq k} \|\hat{\beta} - \beta^*\|_2 \gtrsim \frac{\sigma_{\epsilon}^2}{\sigma_{z}^2} \frac{d^*\log(d/d^*)}{n}.
\]
\end{proof}

\subsection{Proof for Theorem 1}

We first define the Rademacher and Gaussian complexity terms for the representation class $\Phi$. We deliberately use the different complexity notions to differentiate the CL-based and same-structure pre-training. In particular, for CL-based pre-training with $n$ triplets of $(x_i,x_i^+, x_i^-)$, the empirical Rademacher complexity of $\Phi$ is given by:
\[
\Rcal_n(\Phi) = \Ebb_{\vec{\sigma}\in \Rbb^{3d}}\sup_{\phi \in \Phi} \sum_{i=1}^n \big\langle \vec{\sigma}, \big[\phi(x_i),\phi(x_i^+),\phi(x_i^-)\big] \big\rangle,
\]
where $\vec{\sigma}$ is the vector of i.i.d Rademacher random variables. For the same-structure pre-training with $n$ samples of $(x_i,y_i)$, the empirical Gaussian complexity of $\Phi$ is given by:
\[
\Gcal_n(\Phi) = \Ebb_{\vec{\gamma} \in \Rbb^d} \sup_{\phi \in \Phi} \sum_{i=1}^n \big\langle \vec{\gamma}, \phi(x_i) \big\rangle,
\]
where $\vec{\gamma}$ is the vector of  i.i.d Gaussian random variables. Without loss of generality, we assume the loss functions for both CL-based and same-structure pre-training are bounded and $L$-Lipschitz. We first prove the result for the same-structure pre-training.

\begin{proof}
First recall from Section 4 that the downstream classifier is optimized on $n$ sample drawn from $P_{\tau}$ by plugging in $\hat{\phi}$, which we denote by: $f_{\hat{\phi},\, P_{\tau,n}}$. Also, we have defined:
\begin{equation*}
\label{eqn:optimal-risk}
    R^*_{\text{task}} = \min_{\phi \in \Phi} \Ebb_{P_{\tau} \sim \Ecal} \Big[ \min_{f\in\Fcal} \Ebb_{(x,y)\sim P_{\tau}}\ell\big(f\circ\phi(x), y \big) \Big],
\end{equation*}
with $\phi^*$ as the optimum, as well as:
\begin{equation}
\label{eqn:phi-risk}
R_{\text{task}}^*(\hat{\phi}) = \Ebb_{P_{\tau}\sim \Ecal} \Ebb_{(x,y)\sim P_{\tau}} \ell\big(f_{\hat{\phi},\, P_{\tau,n}}(x), y \big).
\end{equation}
Therefore, it holds that:
\begin{equation}
\label{eqn:append-thm-1}
\begin{split}
    R_{\text{task}}(\hat{\phi}) - R^*_{\text{task}} &= R_{\text{task}}(\hat{\phi}) - \frac{1}{n}\sum_i\ell\big(f_{\hat{\phi}, P_{\tau,n}}(x_i), y_i \big)  \\
    & + \frac{1}{n}\sum_i\ell\big(f_{\hat{\phi}, P_{\tau,n}}(x_i), y_i \big) - \frac{1}{n}\sum_i\ell\big(f_{\phi^*, P_{\tau,n}}(x_i), y_i \big)  \\
    & + \frac{1}{n}\sum_i\ell\big(f_{\phi^*, P_{\tau,n}}(x_i), y_i \big) - \Ebb_{(x_i,y_i)\sim P_{\tau,n}} \Big[\frac{1}{n}\sum_i \ell\big(f_{\phi^*, P_{\tau,n}}(x_i), y_i \big) \Big] \\
    & + \Ebb_{(x_i,y_i)\sim P_{\tau,n}} \Big[\frac{1}{n}\sum_i \ell\big(f_{\phi^*, P_{\tau,n}}(x_i), y_i \big) \Big] - \min_{f\in\Fcal} \Ebb_{(X,Y)\sim P_{\tau}} \ell\big(f\circ \phi(X), Y \big).
\end{split}
\end{equation}
We define: $f^* = \arg\min_{f\in\Fcal} \Ebb_{(x,y)\sim P_{\tau}} \ell\big(f\circ \phi(x), y \big)$. Firstly, note that by the definition of $\hat{\phi}$, we have for the second line on RHS of (\ref{eqn:append-thm-1}) that:
\[
\frac{1}{n}\sum_i\ell\big(f_{\hat{\phi}, P_{\tau,n}}(x_i), y_i \big) - \frac{1}{n}\sum_i\ell\big(f_{\phi^*, P_{\tau,n}}(x_i), y_i \big) \leq 0. 
\]
In the next step, notice for the last line on RHS of (\ref{eqn:append-thm-1}) that:
\begin{equation}
\begin{split}
    \Ebb_{(x_i,y_i)\frac{1}{n}\sum_i \sim P_{\tau,n}} \Big[\frac{1}{n}\sum_i \ell\big(f_{\phi^*, P_{\tau,n}}(x_i), y_i \big) \Big] &= \Ebb_{(x_i,y_i)\sim P_{\tau,n}} \min_{f\in\Fcal} \frac{1}{n}\sum_i \ell\big(f\circ\phi^*(x_i), y_i\big) \\
    &\leq \Ebb_{(x_i,y_i)\frac{1}{n}\sum_i \sim P_{\tau,n}} \Big[\frac{1}{n}\sum_i \ell\big(f^*\circ\phi^*(x_i), y_i \big) \Big] \\
    &\leq \min_{f\in\Fcal} \Ebb_{(X,Y)\sim P_{\tau}} \ell\big(f\circ h^*(X), Y \big).
\end{split}
\end{equation}
Henceforth, the last line is also non-positive. As for the third line on RHS of (\ref{eqn:append-thm-1}), notice that is involves a bounded random variable $\frac{1}{n}\sum_i\ell\big(f_{\phi^*, P_{\tau,n}}(x_i), y_i \big)$ (since we assume the loss function is bounded) and its expectation. Using the regular Hoeffding bound, it holds with probability at least $1-\delta$ that:
\[
\frac{1}{n}\sum_i\ell\big(f_{\phi^*, P_{\tau,n}}(x_i), y_i \big) - \Ebb_{(x_i,y_i)\sim P_{\tau,n}} \Big[\frac{1}{n}\sum_i \ell\big(f_{\phi^*, P_{\tau,n}}(x_i), y_i \big) \Big] \lesssim \sqrt{\log(8/\delta)}. 
\]
Therefore, what remains is to bound the first line on RHS of (\ref{eqn:append-thm-1}), which can follows:

\begin{equation}
\begin{split}
    & R_{\text{task}}(\hat{\phi}) - \frac{1}{n}\sum_i\ell\big(f_{\hat{\phi}, P_{\tau,n}}(x_i), y_i \big) \leq \sup_{\phi\in\Phi}\Big\{R_{\text{task}}(\hat{\phi}) - \frac{1}{n}\sum_i\ell\big(f_{\phi, P_{\tau,n}}(x_i), y_i \big) \Big\} \\
    &\leq \sup_{\phi}\Ebb_{P_{\tau}\sim\Ecal} \Ebb_{(x_i,y_i)\sim P_{\tau,n}} \Big[\Ebb_{(X,Y)\sim P_{\tau}} \ell\big(f\circ\phi(X),Y\big) - \frac{1}{n} \sum_i \ell\big( f_{\phi, P_{\tau,n}}(x_i), y_i \big) \Big] \\
    &+ \sup_{\phi\in\Phi} \Big[\frac{1}{n} \sum_i \ell\big( f_{\phi, P_{\tau,n}}(x_i), y_i \big) -  \Ebb_{(x_i,y_i)\sim P_{\tau,n}} \frac{1}{n}\sum_i \ell\big( f_{\phi, P_{\tau,n}}(x_i), y_i \big) \Big].
\end{split}
\end{equation}
Finally, the existing results of bounding empirical processes from Theorem 14 of \cite{maurer2016benefit} shows that with probability at least $1-\delta$, the third line above is bounded by:
\begin{equation*}
\begin{split}
     \frac{\sqrt{2\pi}L\Gcal_n(\Phi)}{\sqrt{n}} + \sqrt{9\log(2/\delta)},
\end{split}
\end{equation*}
and the second line is bounded by:
\[
\frac{\sqrt{2\pi}}{n} Q \sup_{\phi\in\Phi} \Ebb_{(X,Y)\sim P_{\tau}} \|\phi(X)\|_2^2,
\]
where $Q:=\tilde{\Gcal}(\Fcal)$ is some complexity measure of the function class $\Fcal$. By combining the above results, rearranging terms and simplifying the expressions, we obtain the desired result. 
\end{proof}

In what follows, we provide the proof for the CL-based pre-training. 
\begin{proof}
Recall that the risk of a downstream classifier $f$ is given by: $R_{\tau}(f;\phi):= \Ebb_{(X,Y)\sim P_{\tau}} \ell(f\circ\phi(x), y)$, where we let $\ell(\cdot)$ be the widely used logistic loss. When $f$ is a linear model, it induces the loss as: $\ell\big(\theta_1^{\intercal}\phi(x) - \theta_2^{\intercal}\phi(x)\big)$, where $\theta_1,\theta_2$ corresponds to the two classes $y=0$ and $y=1$. We define a particular linear classifier whose class-specific parameters are given by: $\bar{\phi}^{(y)} := \Ebb_{x\sim P_X^{(y)}}\phi(x)$, for $y\in\{0,1\}$. They correspond to using the average item embedding from the same class as the parameter vector. Therefore, we have: 
\[
R_{\tau}(\bar{\phi};\phi):= \Ebb_{(X,Y)\sim P_{\tau}} \ell\big((\bar{\phi}^{(Y)})^{\intercal}\phi(x) - (\bar{\phi}^{(1-Y)})^{\intercal}\phi(x)\big).
\]
The importance of studying this particular downstream classifier is because, as long as $\Fcal$ includes linear model, it holds that: $\min_{f\in\Fcal}R_{\tau}(f;\phi) \leq R_{\tau}(\bar{\phi};\phi)$. Further more, we will be able to derive meaningful results (upper bound) the risk associated with $\bar{\phi}$ with the CL-based pre-training risk. We first define the probability that two randomly drawn instances fall into the same class: $q:= P_Y(y=1)^2 + P_Y(y=0)^2$. In particular, we observe that:
\begin{equation}
\begin{split}
    R_{\text{CL}}(\phi) & = \Ebb_{x,x^+\sim P_{\text{pos}}, x^-\sim P_{\text{neg}}} \big[ \ell\big(\phi(x)^{\intercal}(\phi(x^+) - \phi(x^-)) \big) \big] \\
    & = \Ebb_{y^+,y^-\sim P_Y^2, x\sim P_{X}^{(y^+)}} \Ebb_{x^+\sim P_{X}^{(y^+)}, x^-\sim P_{X}^{(y^-)}} \big[ \ell\big(\phi(x)^{\intercal}(\phi(x^+) - \phi(x^-)) \big) \big] \\
    & \geq  \Ebb_{y^+,y^-\sim P_Y^2, x\sim P_{X}^{(y^+)}} \Big[\ell\Big(\phi(x)\big(\bar{\phi}^{(y^+)} - \bar{\phi}^{(1-y^+)}\big) \Big) \Big] \text{ Jensen's inequality} \\
    & = (1-q)  \Ebb_{y^+,y^-\sim P_Y^2, x\sim P_{X}^{(y^+)}} \Big[\ell\Big(\phi(x)\big(\bar{\phi}^{(y^+)} - \bar{\phi}^{(1-y^+)}\big) \Big) \Big| y^+ \neq y^- \Big] + q \\
    & = (1-q)R_{\tau}(\bar{\phi};\phi) + q.
\end{split}
\end{equation}
Therefore, we conclude the relation between the $\bar{\phi}$-induced classifier and the CL-based pre-training risk: 
\[
R_{\tau}(\bar{\phi};\phi) \leq \frac{1}{1-q}\big(R_{\text{CL}}(\phi) - q \big), \text{for any }\phi\in\Phi.
\]
The next step is to study the generalization bound regarding $R_{\text{CL}}(\phi), \forall \phi \in \Phi$. Suppose the loss function $\ell(\cdot)$ is bounded by $B$, and is $L$-Lipschitz. Both assumptions holds for the logistic loss that we study. We define the CL-specific loss function class on top of $\phi\in\Phi$:
\[
\Hcal_{\Phi} := \Big\{\frac{1}{B}\ell\big(\phi(x)^{\intercal}\big(\phi(x^+) - \phi(x^-) \big) \big) \big| \phi\in\Phi \Big\},
\]
such that for $h_{\phi}\in\Hcal_{\Phi}$ we have: $h_{\phi}(x,x^+,x^-) = \frac{1}{B}\ell\circ \tilde{\phi}(x,x^+,x^-)$, where $\tilde{\phi}$ is the mapping of: $\phi(x),\phi(x^+),\phi(x^-) \mapsto \phi(x)^{\intercal}\big(\phi(x^+) - \phi(x^-)$. The classical generalization result \cite{bartlett2002rademacher} shows that with probability at least $1-\delta$:
\begin{equation}
    \Ebb h_{\phi} \leq \frac{1}{n}\sum_{i=1}^n h_{\phi}(x_i,x_i^+,x_i^-) + \frac{2\Rcal_n(\Hcal_{\Phi})}{n} + 3\sqrt{\frac{\log(4/\delta)}{n}}.
\end{equation}
In what follows, we connect the complexity of $\Rcal_n(\Hcal_{\Phi})$ to the desired $\Rcal_n(\Phi)$. Note that the Jacobian associated with the mapping of $\tilde{\phi}$ is given by: 
\[
J:= \big[\phi(x^+) - \phi(x^-), \phi(x), -\phi(x) \big],
\]
so it holds that $\|J\|_2\leq\|J\|_F\leq 3\sqrt{2}R$, where $R$ is the uniform bound on $\phi\in\Phi$. Hence, $\ell\circ\phi$ is $(3\sqrt{2}LR/B)$-Lipschitz on the domain of $\big(\phi(x), \phi(x^+),\phi(x^-)\big)$. In what follows, using the Telegrand contraction inequality for Rademacher complexity, we reach: $\Rcal_n(\Hcal_{\Phi})\leq 3\sqrt{2}LR/B \Rcal_n(\Phi)$. Combining the above results, we see that for any $\phi\in\Phi$, it holds with probability at least $1-\delta$ that:
\[
R_{\text{CL}}(\phi) \leq \frac{1}{n}\sum_{i=1}^n \ell\big(\phi(x_i)^{\intercal}\big(\phi(x_i^+) - \phi(x_i^-)\big)\big) + \Ocal\Big(R\Rcal(\Phi) + \sqrt{\frac{\log(4/\delta)}{n}} \Big).
\]
Finally, since we have the decomposition: $R_{\text{CL}}(\phi) = R_{\text{CL}}^G(\phi) + R_{\text{CL}}^B(\phi)$, it remains to bound: $R_{\text{CL}}^B(\phi) = \Ebb_{y}\Ebb_{x,x+,x^-\sim P_X^{y}} \Big[\ell \big( \phi(x)^{\intercal}\big(\phi(x^+) - \phi(x^-) \big) \Big]$. 

Let $z_i := \phi(x_i)^{\intercal}\big(\phi(x_i^+) - \phi(x_i^-)\big)$ and $z = \max z_i$. It is straightforward to show for logistic loss that: $R_{\text{CL}}^B(\phi)\leq \Ebb|z|$. Further more, we have:
\begin{equation}
\begin{split}
    E|z| \leq \Ebb\big[\max_i |z_i| \big] & \leq n\Ebb[|z_1|] \\
    & \leq n\Ebb_x\Big[\|\phi(x)\|\sqrt{\Ebb_{x^+,x^-}\Big(\phi(x)/\|\phi(x)\| \big(\phi(x^+) - \phi(x^-)\big)\Big)^2} \Big] \\ 
    & \lesssim R\Ebb_y\big\|\text{cov}_{P_X^{(y)}}\phi \big\|_2.
\end{split}
\end{equation}
Henceforth, $R_{\text{CL}}(\phi) \lesssim R^G_{\text{CL}}(\phi) + R\Ebb_y\big\|\text{cov}_{P_X^{(y)}}\phi \big\|_2$. Recall that $R^*_{\text{task}} = \min_{\phi}R^G_{CL}(\phi)$ and for all $\phi\in\Phi$, we have $R_{\text{task}}(\phi) \leq \min_{f\in\Fcal} R_{\tau}(f;\phi) \leq R_{\tau}(\hat{\phi};\phi)$. Hence, by rearranging terms and discarding constant factors, we reach the final result:
\begin{equation*}
    R_{\text{task}}(\hat{\phi}) - R^*_{\text{task}} \lesssim \frac{\Gcal_n(\Phi)}{\sqrt{n}} + \frac{R\tilde{\Gcal}(\Fcal)}{n} + \sqrt{\log(8/\delta)},
\end{equation*}
\end{proof}

\subsection{Proof for Proposition 2}

\begin{proof}
Recall that the kernel-based classifier is given by:
\[
f_{\phi}(x) = \frac{E_{x'}\big[y'k_{\phi}(x,x') \big]}{\sqrt{\Ebb[k_{\phi}^2]}},
\]
where $y\in\{-1,+1\}$ and $R^{\text{OOD}}$ is the out-of-distribution risk associated with a $0-1$ classification risk. We first define for $x\in\Xcal$:
\[
\gamma_{\phi}(x):= \sqrt{\frac{\Ebb_{x'}\big[ K_{\phi}(x, x')\big]}{\Ebb_{x,x'}\big[K_{\phi}(x,x') \big]}},
\]
where the expectation is taken wrt. the underlying distribution. Using the Markov inequality, we immediately have: $|\gamma(x)| \leq \frac{1}{\sqrt{\delta}}$ with probability at least $1-\delta$. It then holds that:
\begin{equation*}
\begin{split}
    1 - R^{\text{OOD}}(f_{\phi}) &= P\big(yf_{\phi}(x) \geq 0\big) \\
    & \geq \Ebb\Big[\frac{yf_{\phi}(x)}{\gamma(x)}\cdot 1[yf_{\phi}(x) \geq 0]  \Big] \\
    &\geq \Ebb\Big[\frac{yf_{\phi}(x)}{\gamma(x)}\Big] \geq \frac{\Ebb\big[K_Y(y,y')K_{\phi}(x,x') \big]}{\sqrt{\Ebb K_{\phi}^2}}\sqrt{\delta} \text{ ,with probability }1-\delta,
\end{split}
\end{equation*}
where $K_Y(y,y') = 1[y=y']$. It concludes the proof.
\end{proof}

\section{Proof for Proposition 3}

\begin{proof}
Recall that the sequential interaction model is given by:
\begin{equation}
    \label{eqn:prop3-1}
    p\big(x_{k+1} \,|\, s \big) = \lambda p_0(x_{k+1}) + (1-\lambda) \frac{\exp\big(\big\langle \phi(x_{k+1}), \varphi(s) \big\rangle \big)}{Z_s}, \, \lambda \in (0,1),
\end{equation}
so the likelihood of the sequence $\{x_1,\ldots,x_{k+1}\}$ is given by:
\[
\prod_{i=1}^{k+1} \Big( \lambda p_0(x_{i}) + (1-\lambda) \frac{\exp\big(\big\langle \phi(x_{i}), \varphi(s) \big\rangle \big)}{Z_s} \Big).
\]
As a result, the log-likelihood of the sequence embedding $\varphi(s)$, for a particular $x_i$ is given by:
\[
l_i\big(\varphi(s)\big) = \log\Big( \lambda p_0(x_{i}) + (1-\lambda) \frac{\exp\big(\big\langle \phi(x_{i}), \varphi(s) \big\rangle \big)}{Z_s} \Big),
\]
and by Taylor approximation, we immediately have:
\begin{equation}
\begin{split}
    f_i(\varphi(s)) = \frac{1-\lambda}{\lambda Z_s p_0(x_i) + (1-\lambda)}\big\langle \phi(x_i),\varphi(s) \big\rangle + f_i(\mathbf{0}) + \text{residual}.
\end{split}
\end{equation}
Note that: $\arg\max_{v:\|v\|_2=1} \langle v, \phi(x_i) \rangle = \phi(x_i)/\|\phi(x_i)\|_2$, so putting aside the residual terms, the approximate optimal achieved is given by:
\[
\arg\max_{\varphi(s)} \sum_{i=1}^{k+1} \Big(\frac{1-\lambda}{\lambda Z_s p_0(x_i) + (1-\lambda)}\big\langle \phi(x_i),\varphi(s) \big\rangle\Big) \propto \sum_{i=1}^k \frac{\alpha}{p_0(x_i)+\alpha} \phi(x_i),
\]
where $\alpha=(1-\lambda) / (\lambda Z_s)$. This concludes the proof.
\end{proof}

\bibliographystyle{abbrvnat}
\bibliography{references}